\documentclass[10pt,letterpaper]{article}
\usepackage{arxiv}
\usepackage{preprint}

\usepackage{amsmath,amssymb}
\usepackage{graphicx}
\usepackage[table,dvipsnames]{xcolor}
\usepackage{booktabs,array,tabularx,colortbl}
\usepackage{caption}
\usepackage[numbers,sort&compress]{natbib}
\usepackage{chapterbib}
\usepackage[shortlabels,inline]{enumitem}

\usepackage[T1]{fontenc}
\usepackage{makecell}
\usepackage{multirow}
\usepackage{threeparttable}
\usepackage{placeins}
\usepackage{microtype}
\definecolor{PAHeader}{HTML}{F2F5F7}
\definecolor{PABlue}{HTML}{205C91}
\providecommand{\PABest}[1]{\textcolor{PABlue}{\textbf{#1}}}

\providecommand{\PAbench}[2]{\mbox{#1~\citep{#2}}}
\providecommand{\PAyes}{\ensuremath{\checkmark}}
\providecommand{\PAno}{\ensuremath{\times}}
\providecommand{\PApart}{\ensuremath{\triangle}}

\usepackage{fontawesome5}
\usepackage{hyperref}
\hypersetup{
  colorlinks=true,linkcolor=PABlue,citecolor=PABlue,urlcolor=PABlue,
  pdftitle={PhysAlign: A Benchmark for Evidence-Grounded Role Alignment in Multimodal Physics Reasoning},
  pdfauthor={Kecheng Liang, Haoyang Liu, Zexin Chen, Zirong Liu, Weixing Chen, Qiufeng Wang, Yang Liu}
}
\newcommand{\PAresource}[3]{%
  \href{#2}{#1\hspace{0.45em}\nolinkurl{#3}}%
}
\renewcommand{\shorttitle}{PhysAlign}
\title{PhysAlign: A Benchmark for Evidence-Grounded\\Role Alignment in Multimodal Physics Reasoning}
\author{%
  \textbf{Kecheng~Liang}\textsuperscript{1,}\thanks{Equal contribution.}\quad
  \textbf{Haoyang~Liu}\textsuperscript{2,}\footnotemark[1]\quad
  \textbf{Zexin~Chen}\textsuperscript{1}\quad
  \textbf{Zirong~Liu}\textsuperscript{1}\\[5pt]
  \textbf{Weixing~Chen}\textsuperscript{1}\quad
  \textbf{Qiufeng~Wang}\textsuperscript{3}\quad
  \textbf{Yang~Liu}\textsuperscript{1,}\thanks{Corresponding author: \href{mailto:liuy856@mail.sysu.edu.cn}{\nolinkurl{liuy856@mail.sysu.edu.cn}}.}\\[8pt]
  {\normalfont\textsuperscript{1}Sun Yat-sen University}\\[2pt]
  {\normalfont\textsuperscript{2}Dalian University of Technology}\\[2pt]
  {\normalfont\textsuperscript{3}Xi'an Jiaotong-Liverpool University}
}
\date{}
\begin{document}
\hypersetup{urlcolor=black}
\maketitle
\setcounter{footnote}{0}
\begin{center}
\small
\begin{tabular}{@{}c@{}}
\PAresource{\faDatabase}{https://huggingface.co/datasets/Jetson888/PhysAlign}{huggingface.co/datasets/Jetson888/PhysAlign}\\[2pt]
\PAresource{\faGlobe}{https://physalign-lab.github.io/}{physalign-lab.github.io/}\\[2pt]
\PAresource{\faGithub}{https://github.com/HCPLab-SYSU/PhysAlign}{github.com/HCPLab-SYSU/PhysAlign}
\end{tabular}
\end{center}
\hypersetup{urlcolor=PABlue}
\vspace{0.35em}
\begin{abstract}
A key challenge in physics diagram understanding is correctly associating visual information with the physical entities, relations, and conditions it describes. Even when a value, symbol, or other local element is accurately recognized, assigning it to the wrong entity or scope can distort the underlying physical premise and lead to incorrect reasoning. To systematically study this challenge, we introduce \textbf{PhysAlign}, a benchmark designed to assess whether multimodal models correctly associate information recognized from physics diagrams with its intended physical role. By disentangling visual recognition from physical-role assignment through localized probes and controlled variants, PhysAlign isolates correspondence errors from recognition failures. It contains 3,341 human-validated probes spanning 986 physics problems, enabling systematic evaluation of visual recognition and physical-role correspondence at scale. We further introduce five complementary evaluation metrics, including CAcc, GAcc, and JAcc, which provide a comprehensive assessment of models' ability to recognize diagram content, establish correct physical correspondences, and solve the underlying physics problem.
Across ten multimodal models, grounding error rates conditional on correct local recognition range from 13.8\% to 60.3\%.
Supplying ground-truth local readings yields paired grounding gains of 1.23--9.31 percentage points, yet errors persist in every model.
These results show that accurate local recognition does not ensure correct physical-role grounding and motivate evaluating evidence--role correspondence alongside final-answer accuracy.
\end{abstract}
\section{Introduction}
\label{sec:intro}

Physics reasoning tests whether multimodal large language models (MLLMs) can
interpret a physical situation from text and diagrams and apply the appropriate
physical laws. A diagram specifies relations among objects and associates
quantities with particular entities, while the accompanying text provides
further conditions and constraints. These correspondences determine the
physical premises on which a solution rests. Evaluating physics reasoning
therefore requires examining whether a model recovers the stated problem setup
as well as whether it derives the correct answer.

Recent benchmarks examine both the use of visual evidence and intermediate
steps in physics reasoning. SeePhys~\citep{xiang2026seephys} distinguishes
problems that require visual information from those that can be solved from
text alone. PhysicsArena~\citep{dai2025physicsarena} evaluates variable
identification, physical process formulation, and solution derivation, while
SeePhys Pro~\citep{xiang2026seephyspro} studies visual variable grounding
through semantically aligned input variants that transfer information from
text to images. However, diagnosing errors in the recovered problem setup
requires a more localized distinction between reading content and assigning it
to the correct physical role. Final-answer accuracy cannot determine whether
an error originates in recognition, role assignment, or subsequent reasoning.
Reconstructing a complete physical graph can also entangle an incorrect
correspondence with unrelated entity omissions, naming differences, and
structured-output errors. We focus on measuring recognition and role
assignment separately at the same evidence location.

\begin{figure*}[t]
    \centering

    \includegraphics[
        width=\textwidth
    ]{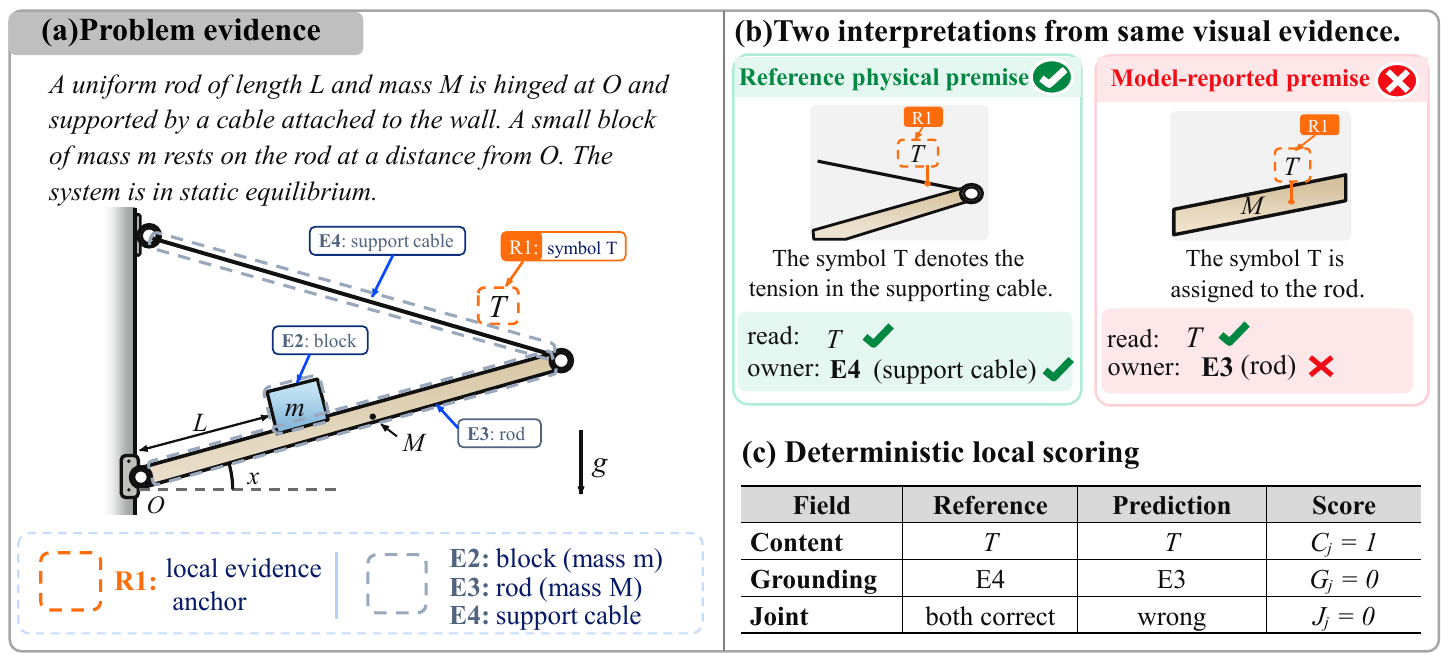}

    \captionsetup{skip=3pt}
    \caption{
        \textbf{Correct recognition does not guarantee
        physical-role grounding.}
        (a) The problem evidence specifies a rod supported by a cable,
        with \texttt{R1} marking the symbol $T$.
        (b) The reference associates $T$ with the supporting cable
        (\texttt{E4}), whereas the illustrated response assigns it to the rod
        (\texttt{E3}).
        (c) Recognition is correct ($C_j = 1$), but grounding and joint
        correctness fail ($G_j = 0$, $J_j = 0$).
        Reference targets and scores are used only for evaluation and are not
        provided to the model.
    }

\label{fig:example}

\end{figure*}

Correct recognition of a symbol does not necessarily imply correct grounding to the corresponding physical entity.
As illustrated in Figure~\ref{fig:example}, a rod carrying a block is held in
static equilibrium by a hinge at $O$ and a supporting cable. The symbol $T$ at
evidence anchor \texttt{R1} denotes the tension in the supporting cable
(\texttt{E4}). The illustrated response recognizes $T$ correctly but assigns it to
the rod (\texttt{E3}). Here, the queried correspondence refers to the cable associated with the tension symbol, rather than to the body on which the tension force acts. We
call the assignment of local content to the physical entity or role supported
by the problem evidence \emph{physical-role grounding}. This motivates
testing whether role-assignment errors persist when evidence locations are
given, even when correct local readings are supplied.


We introduce \textbf{PhysAlign}, a benchmark for diagnosing physical-role
grounding through localized probes with explicit evidence anchors. PhysAlign
contains 3,341 probes constructed from 986 physics problems across six source
datasets. We convert human-reviewed observational annotations into tests of
entity reference and physical-quantity ownership. Each probe retains the
original problem statement, diagram, and physical conditions, and uses an
image region or text span to identify the queried evidence. For each probe, the model identifies the physical role associated with the queried content and, when an independent reading target is available, also reports the corresponding local content. This design enables recognition and role assignment to be evaluated against their respective evidence without requiring full diagram-graph reconstruction.

PhysAlign combines separate recognition and grounding scores with a paired
reading control. For probes with independently defined reading and role
targets, we use deterministic rules to score recognition and grounding
separately, joint correctness requires both targets to be correct. These scores
distinguish correctly recognized content from a correctly reported physical
premise. A subset of 553 probes also supports paired input variants: the
original problem, evidence anchor, question, candidates, and scoring rules
remain fixed, while the augmented condition supplies predefined correct local
readings. We select these readings using model-independent rules that prohibit
disclosure of the queried role correspondence. The comparison measures whether
supplying correct content improves grounding. This decomposition concerns
observable responses and does not assume separate recognition and grounding
modules within a model.

Results from ten MLLMs show that grounding errors persist across model
families and scales even when local content is recognized correctly.
Grounding error rates conditional on correct recognition range from
13.8\% to 60.3\%. Supplying reference readings yields paired grounding
gains of 1.23--9.31 percentage points on model-specific observed subsets,
with the highest +GT accuracy reaching 84.90\%.
Item-level analysis of the three Qwen3.5 models and InternVL3.5-8B~\citep{wang2025internvl3}
reveals both repairs and harms, so positive aggregate changes do not
imply uniform benefits.
These findings support evaluating reported physical premises against
their source evidence alongside original-problem solving, without
treating supplied readings as perfect perception.

\noindent\textbf{Our contributions are threefold:}
\begin{enumerate}
    \setlength{\topsep}{2pt}
    \setlength{\itemsep}{1pt}
    \setlength{\parsep}{0pt}

    \item \textbf{Benchmark.}
    We introduce \textbf{PhysAlign}, a benchmark for local
    physical-role grounding, with evidence-anchored probes
    of entity reference and physical-quantity ownership
    and human-reviewed role targets.

    \item \textbf{Evaluation protocol.}
    We define deterministic recognition, grounding, and
    joint scores, together with a paired reading control
    that supplies correct local content without revealing
    the target role correspondence.

    \item \textbf{Empirical findings.}
    Across model families and scales, we identify cases
    where correct recognition coexists with incorrect
    physical-role assignments, including probes where
    a frozen proximity rule fails. Paired readings reveal
    both repairs and harms, exposing errors that
    final-answer accuracy alone cannot localize.
\end{enumerate}

\section{Related Work}
\label{sec:related_work}

\paragraph{Multimodal scientific and physics reasoning.}
Scientific reasoning benchmarks span science question answering in
ScienceQA~\citep{lu2022learn}, multidisciplinary reasoning in
MMMU~\citep{yue2024mmmu}, visual mathematics in
MathVista~\citep{lu2024mathvista}, and undergraduate physics in
PhysUniBench~\citep{wang2025physunibench}.
SeePhys~\citep{xiang2026seephys} distinguishes
\emph{Vision-Essential} from \emph{Vision-Optional} problems,
while SeePhys Pro~\citep{xiang2026seephyspro} studies robustness
as task-critical information moves from text to diagrams.
PhysicsArena~\citep{dai2025physicsarena} further evaluates
intermediate stages such as variable identification and process
construction.
These benchmarks primarily assess final solutions or broad reasoning
stages, where perception, entity correspondence, and derivation
remain coupled.
PhysAlign instead examines the pre-reasoning recovery of the problem
setup: whether content read from a local visual region is assigned
to the correct physical entity or role.

\paragraph{Controlled visual diagnosis and role grounding.}
Controlled vision--language evaluations show that recognizing individual
elements does not ensure correct relational interpretation.
Winoground~\citep{thrush2022winoground},
ARO~\citep{yuksekgonul2023aro},
VL-CheckList~\citep{zhao2022vlchecklist}, and
SugarCrepe~\citep{hsieh2023sugarcrepe} isolate failures in compositional
structure, attributes, relations, and order. Related diagnostics for
mathematical and scientific reasoning vary the availability or modality
of visual evidence: MathVerse~\citep{zhang2024mathverse} changes the
visual--textual composition of matched problems,
MMStar~\citep{chen2024we} emphasizes vision-essential evaluation and
textual-shortcut control, and
SeePhys Pro~\citep{xiang2026seephyspro} probes modality transfer and
visual variable grounding. PhysAlign instead fixes the original problem
and evidence anchors, separately measures \emph{Content Recognition}
and \emph{Role Grounding}, and uses paired controls that supply correct
local readings without revealing the grounding target.
Benchmark protocols are compared in
Appendix~\ref{app:benchmark_comparison}
(Table~\ref{tab:related_benchmarks}).


\begin{figure*}[!t]
    \centering
    \includegraphics[width=0.87\textwidth]{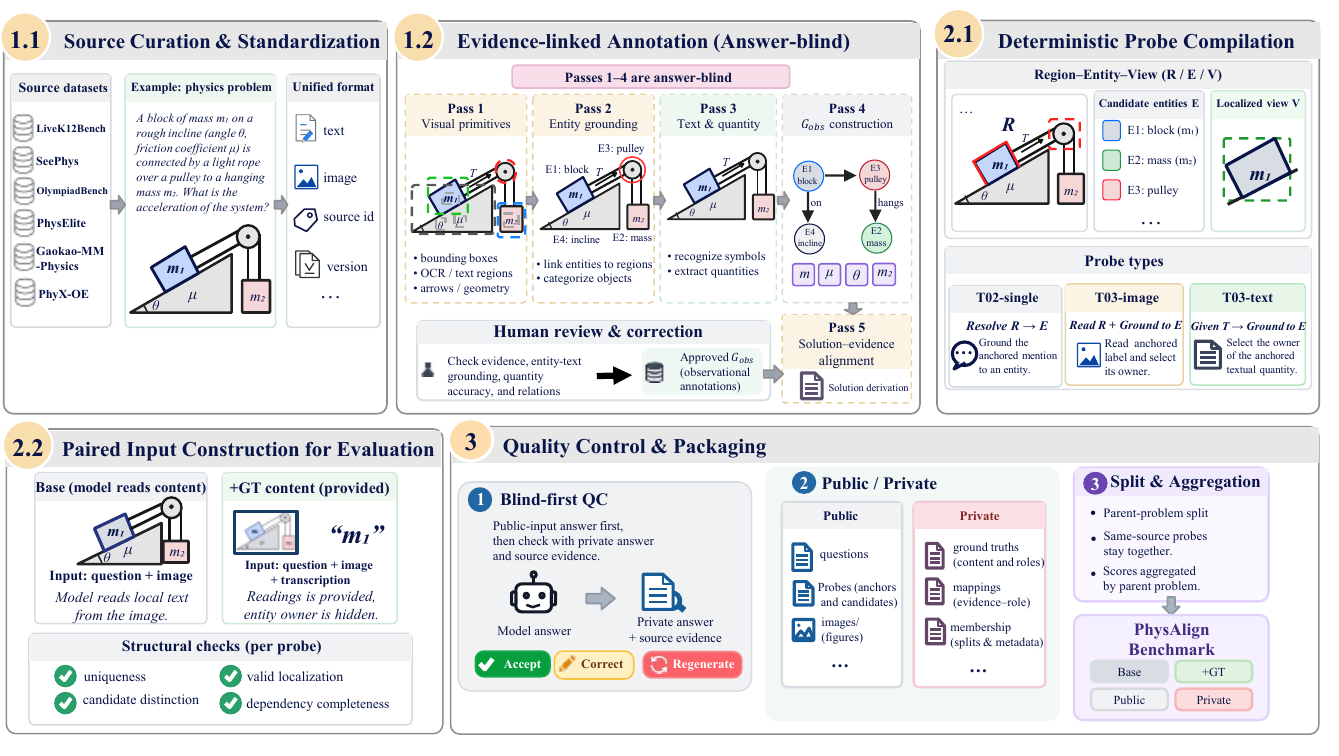}

\caption{
        \textbf{Data construction and quality assurance in PhysAlign.}
        Standardized source problems undergo answer-blind annotation
        (Passes~1--4), human review, and probe compilation.
        Pass~5 aligns reference solutions with evidence for analysis
        only and does not supply probe targets.
        Paired base/+GT inputs share the original problem, evidence
        anchors, questions, and candidates. +GT supplies correct
        local readings without disclosing the queried role target.
        Structural checks and instance audits precede packaging,
        which separates public inputs from private targets and
        scoring mappings.
    }

\label{fig:construction}

\end{figure*}

\section{PhysAlign: An Evidence-Grounded Role Alignment Benchmark}
\label{sec:benchmark}

\subsection{Task Definition and Scope}
\label{sec:task}

Each PhysAlign probe evaluates a local physical correspondence using
the original problem text and diagram, an evidence anchor, a local
question, and candidates identified by neutral aliases. The anchor
specifies an image region or an exact text span. The model selects
the corresponding physical role target and, when a separate reading
target is defined, reports the content at the anchor.

In Figure~\ref{fig:example}, anchor \texttt{R1} marks the tension symbol
$T$, and the queried correspondence is the supporting cable
(\texttt{E4}). A response with \texttt{read}: $T$ and
\texttt{owner}: \texttt{E3} correctly recognizes the symbol but
incorrectly assigns it to the rod. For this query, the target is
the cable associated with the tension, rather than the body on
which the force acts.

The benchmark uses three probe types:
\texttt{T02-single} resolves an anchored mention to an entity,
\texttt{T03-image} transcribes an anchored quantity label and
identifies its owner, and \texttt{T03-text} identifies the owner of
an anchored textual quantity. Each query preserves the relation
specified by its source annotation, and a quantity's owner may be
an object, force, or field. Joint evaluation requires separately
defined targets for content recognition and role grounding.
Probes without an independent reading target receive grounding
scores only. This decomposition concerns observable responses
and makes no assumption about separate recognition and grounding
modules within a model.

\subsection{Evidence-Linked Annotations and Probe Construction}
\label{sec:probe}

We construct probes from reviewed annotations linked to the
original problem evidence. Figure~\ref{fig:construction} summarizes
the pipeline, and Appendix~\ref{app:dataset} provides provenance
records and admission criteria.

We use GPT-5.6 Sol~\citep{openai2026gpt56sol} with high reasoning
effort to draft structured annotations,
which are then reviewed and corrected against the original text
and diagrams. Passes~1--4 record visible primitives, entity
references, explicit quantities and conditions, and their supporting
evidence without access to reference answers or solutions.
Only these reviewed observational records supply probe targets.
Pass~5 separately aligns reference solutions with evidence
for analysis.

Fixed templates convert supported annotation fields into local
questions. Candidates use neutral aliases linked to image regions
or text spans. Alias assignments and candidate order are randomized
once and held fixed across models. Single-target probes require
at least two valid, type-compatible candidates, set-valued templates
retain all correct targets. Candidate descriptions exclude
answer-bearing role associations. Reference readings, role targets,
alias mappings, and source-annotation links are stored separately
from public inputs. Eligible probes also receive the paired +GT
input defined in Section~\ref{sec:paired_conditions}.

Quality assurance comprises source-annotation review,
model-assisted probe screening, and human verification of exported
probes. Checks assess evidence localization, question semantics,
candidate completeness and distinguishability, target support,
and information leakage. During human verification, two coauthors
independently answer the public probe before consulting private
targets and source annotations, a third reviewer adjudicates
disagreements. Ambiguous or unsupported records are withheld
pending resolution. Appendix~\ref{app:review} details construction-stage verification.

\noindent\textbf{Post-construction audit.}
We audited 2,400 of the 3,341 retained probes (71.83\%), sampling by
source group independently of model scores. We inspected every probe in
the three smallest groups and sampled without replacement in the three
largest. Of the audited probes, 2,332 (97.17\%)
passed without revision. The 68 flagged probes had issues with local
transcriptions, evidence locations, ownership labels, or candidate
mappings. After correction, 66 passed reinspection. The two unresolved
probes were excluded from the subsequent release candidate
(Appendix~\ref{app:quality_audit}).

\begin{table*}[t]
    \centering
    \caption{Source coverage and release partition of PhysAlign.}
    \label{tab:dataset_statistics}

    \begingroup
    \scriptsize
    \setlength{\tabcolsep}{2pt}
    \renewcommand{\arraystretch}{1}

    \begin{minipage}[t]{0.52\textwidth}
        \vspace{0pt}
        \raggedright
        \textbf{(a) Source coverage}\par
        \vspace{3pt}

        \begin{tabular*}{\linewidth}{@{\extracolsep{\fill}}lrr@{}}
            \toprule
            \textbf{Source dataset}
            & \textbf{Parents}
            & \shortstack[r]{\textbf{Parents}\\\textbf{with +GT}} \\
            \midrule
            SeePhys~\citep{xiang2026seephys}
            & 268 & 119 \\
            LiveK12Bench~\citep{wang2026livek12bench}
            & 250 & 88 \\
            PhysElite~\citep{xu2026physelite}
            & 220 & 79 \\
            OlympiadBench\textsuperscript{a}~\citep{he2024olympiadbench}
            & 209 & 68 \\
            Gaokao-MM-Physics~\citep{zong2024gaokao}
            & 23 & 17 \\
            PhyX-OE~\citep{shen2025phyx}
            & 16 & 14 \\
            \midrule
            \textbf{Full release}
            & \textbf{986} & \textbf{385} \\
            \bottomrule
        \end{tabular*}

        \par\vspace{4pt}
        {\footnotesize
            \textsuperscript{a} English physics competition subset.\par
        }
    \end{minipage}\hfill
    \begin{minipage}[t]{0.45\textwidth}
        \vspace{0pt}
        \raggedright
        \textbf{(b) Development/test partition}\par
        \vspace{3pt}

        \begin{tabular*}{\linewidth}{@{\extracolsep{\fill}}lrrr@{}}
            \toprule
            \textbf{Split}
            & \textbf{Parents}
            & \shortstack[r]{\textbf{Local}\\\textbf{probes}}
            & \shortstack[r]{\textbf{Paired}\\\textbf{+GT}} \\
            \midrule
            Development & 79 & 208 & 90 \\
            Test & 907 & 3,133 & 463 \\
            \midrule
            \textbf{Full release}
            & \textbf{986} & \textbf{3,341} & \textbf{553} \\
            \bottomrule
        \end{tabular*}

        \par\vspace{4pt}
        {\footnotesize
            \textit{Evaluation.} Both splits are used jointly.
            Paired +GT inputs are variants, not additional unique probes.\par
        }
    \end{minipage}
    \par
    \endgroup
\end{table*}

\subsection{Dataset Scale and Coverage}
\label{sec:quality_dataset}

PhysAlign contains \textbf{986 parent problems} from six sources and
\textbf{3,341 unique localized probes}, \textbf{553 probes} from 385
parents also support paired +GT inputs, without adding unique probes.
Table~\ref{tab:dataset_statistics} reports source coverage and the
parent-disjoint release partition. Evaluation pools are drawn from the full
release (development and test combined), metric-specific eligibility and
observed coverage are defined in Section~\ref{sec:scoring}.

\begin{figure*}[!t]
    \centering
    \includegraphics[width=0.95\textwidth]{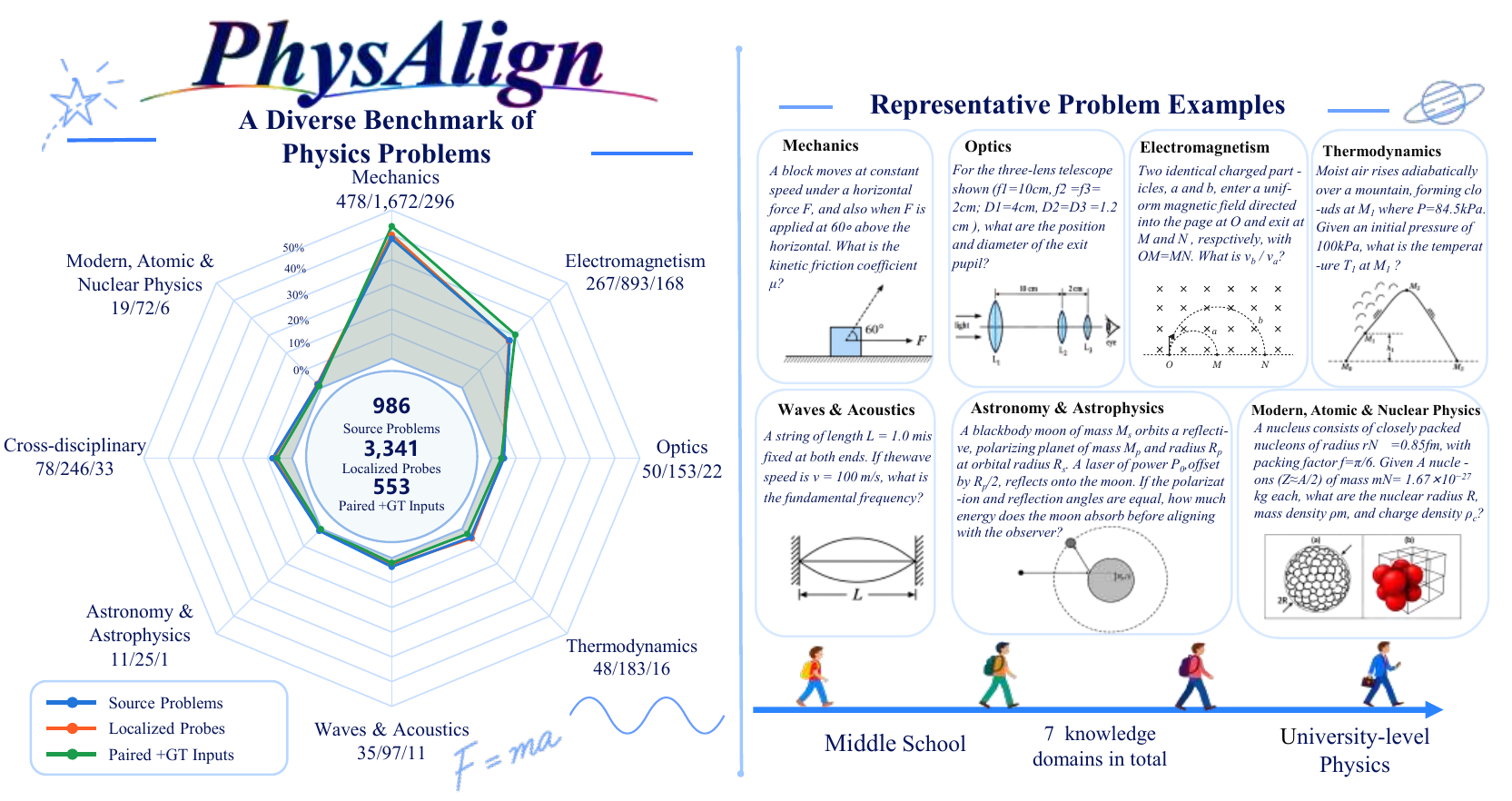}
    
    \caption{%
        \textbf{Domain coverage and representative physics problems in PhysAlign.}
        \textbf{Left:} Distributions of source problems, localized probes,
        and paired +GT inputs across seven physics domains and a
        cross-disciplinary group.
        Each radar series shows percentages within its corresponding
        collection, counts beside each group follow the same order.
        \textbf{Right:} Representative question--diagram examples from
        the seven domains, illustrating an educational span from
        middle school to university-level physics.
    }
    \label{fig:dataset-representation}

\end{figure*}

The source problems span mechanics, electromagnetism, optics,
thermal physics, waves, modern physics, and astrophysics
(Figure~\ref{fig:dataset-representation}). Mechanics and
electromagnetism account for most parent problems under the
estimated domain assignments. The illustrated problems provide
context for the local correspondences evaluated by PhysAlign.
Appendix~\ref{app:dataset_statistics} reports estimated domain
composition, probe density, and paired coverage, with parent-level
and probe-level statistics reported separately.

\section{Evaluation Protocol and Metrics}
\label{sec:paired}

\subsection{Paired Evaluation with Ground-Truth Local Content}
\label{sec:paired_conditions}

For each eligible probe, the \textit{Base} input contains the original problem,
local question, evidence anchors, and candidates. The matched \textit{+GT}
input additionally supplies verified ground-truth (GT) readings of the specified
local content while withholding the queried physical role. Both conditions use
the same question, candidates and aliases, output schema, inference settings,
and scoring rules, and are run in independent sessions without shared history,
predictions, or feedback. Supplied readings are inputs and are therefore
excluded from recognition scoring.

Model-independent rules determine which readings may be supplied. In
Figure~\ref{fig:example}, +GT may state that R1 reads $T$, but not that
$T$ denotes the tension in the supporting cable (E4), since the latter reveals
the grounding target. Candidate-side information, when present, is selected
symmetrically across candidates. The paired comparison measures how grounding
changes when verified local content is supplied. It does not provide a complete
physical description or isolate the causal effect of correcting OCR errors.

\subsection{Deterministic Scoring and Balanced Aggregation}
\label{sec:scoring}

Let $\mathcal G$ contain all 3,341 grounding probes from 986 parents.
The joint set $\mathcal L\subseteq\mathcal G$ contains 412 probes from
311 parents with independent reading targets. The paired-eligible release
set $\mathcal P\subseteq\mathcal G$ contains 553 probes from 385 parents
with valid +GT inputs. Recognition and joint metrics use $\mathcal L$,
and overall grounding uses $\mathcal G$. Base/+GT comparisons use model
$m$'s \texttt{T03-image} probes with both response records available,
$\mathcal P_m\subseteq\mathcal L\cap\mathcal P$
(Appendix~\ref{app:paired_support}).

For predicted content $\hat{c}_j$ and candidate identifier $\hat{u}_j$, let
$c_j$ and $r_j$ denote the reference content and role target. Given the fixed
content normalizer $N_j$ and candidate mapping $\phi_j$,
\begin{equation}
\begin{aligned}
C_j &= \mathbf{1}[N_j(\hat{c}_j) = N_j(c_j)],\\
G_j &= \mathbf{1}[\phi_j(\hat{u}_j) = r_j],\qquad J_j = C_jG_j.
\end{aligned}
\label{eq:item_scores}
\end{equation}
$C_j$ and $J_j$ are defined on $\mathcal{L}$, whereas $G_j$ applies to
$\mathcal{G}$. Content must match the specified anchor rather than an unordered
inventory. Grounding is scored independently, while $J_j$ requires correct
recognition and grounding for the same local premise. Base and +GT use the same
grounding rule. Normalization, multi-field and set matching, and format-error
handling are fixed before evaluation, these local scores use no
open-ended language-model judge.

For an evaluation set $\mathcal{S}$, $\operatorname{Agg}_{\mathcal{S}}$ first
averages probes within each parent--task-type group, then parents within each
task type, and finally task types using prespecified weights
(see Appendix~\ref{app:scoring_details} for the formal definition).
We report
\begin{equation}
\begin{aligned}
CAcc &= \operatorname{Agg}_{\mathcal L}(C),\qquad
JAcc = \operatorname{Agg}_{\mathcal L}(J),\\
GAcc_{\mathcal S}^{v} &= \operatorname{Agg}_{\mathcal S}(G^v),
\quad v\in\{\mathrm{base},+\mathrm{GT}\}.
\end{aligned}
\label{eq:aggregate_scores}
\end{equation}
This aggregation prevents annotation-rich parents from dominating a task type.
Overall grounding is $GAcc_{\mathcal{G}}^{\mathrm{base}}$.
Paired scores use identical Base/+GT support and weights within each
$\mathcal{P}_m$, with
$\Delta G_{\mathcal{P}_m}=GAcc_{\mathcal{P}_m}^{+\mathrm{GT}}-
GAcc_{\mathcal{P}_m}^{\mathrm{base}}$, +GT performance is not assumed to
be an upper bound.

On $\mathcal{L}$, $C^{+}G^{-}$ denotes $C_j=1,G_j=0$, i.e., correct
content recognition with incorrect grounding, other sign combinations are
defined analogously. Under the same aggregation, its weighted share is
$CAcc-JAcc$, not $CAcc-GAcc$, $JAcc$ is not the product of marginal
accuracies. We report $GAcc_{\mathcal{L}}^{\mathrm{base}}$ for same-set
comparisons.

\begin{table*}[t]
\centering
\caption{Recognition, grounding, paired local-reading controls, and independent original-problem solving.}
\label{tab:main_results}
\begingroup
\small
\setlength{\tabcolsep}{3.2pt}
\renewcommand{\arraystretch}{1}
\begin{tabular*}{\linewidth}{@{\extracolsep{\fill}}lrrrrrrrr@{}}
\toprule
& \multicolumn{3}{c}{\textbf{Joint} ($\mathcal L$)}
& \multicolumn{1}{c}{\textbf{All} ($\mathcal G$)}
& \multicolumn{3}{c}{\textbf{Reading control} ($\mathcal P_m$)}
& \multicolumn{1}{c}{\textbf{Solving}}\\
\cmidrule(lr){2-4}\cmidrule(lr){5-5}\cmidrule(lr){6-8}\cmidrule(l){9-9}
\rowcolor{PAHeader}
\textbf{Model} & $CAcc\,\uparrow$ & $GAcc^{\mathrm{base}}_{\mathcal L}\,\uparrow$ & $JAcc\,\uparrow$ & $GAcc\,\uparrow$ & $GAcc^{\mathrm{base}}_{\mathcal P_m}\,\uparrow$ & $GAcc^{+\mathrm{GT}}_{\mathcal P_m}\,\uparrow$ & $\Delta G$ & $SolveAcc\,\uparrow$ \\
\midrule
Qwen3.5-4B~\citep{qwen35blog} & 53.97 & 46.51 & 32.09 & 40.21 & 46.51 & 54.61 & +8.10 & 9.94 \\
Qwen3.5-9B~\citep{qwen35blog} & 65.81 & 54.55 & 37.55 & 40.52 & 54.55 & 56.53 & +1.98 & 12.12 \\
Qwen3.5-27B~\citep{qwen35blog} & 84.14 & 80.29 & 68.12 & 84.11 & 80.29 & 82.00 & +1.71 & 18.39 \\
InternVL3.5-8B~\citep{wang2025internvl3} & 2.12 & 33.36 & 1.05 & 23.52 & 33.36 & 34.81 & +1.45 & 7.47 \\
GPT-6 Astra~\citep{openai2026gpt6astra} & \PABest{89.44} & \PABest{83.31} & \PABest{77.13} & 81.94 & \PABest{83.59} & \PABest{84.90} & +1.32 & \PABest{71.72} \\
Gemini-3.8-flash~\citep{googledeepmind2026gemini38flash} & 85.20 & 79.33 & 70.59 & \PABest{90.71} & 78.67 & 80.86 & +2.20 & 21.31 \\
Gemma4-26B-A4B~\citep{team2026gemma} & 66.45 & 63.87 & 47.38 & 69.27 & 63.87 & 65.10 & +1.23 & 0.23 \\
GLM46V-flash~\citep{vteam2025glm45vglm41vthinkingversatilemultimodal} & 26.86 & 51.23 & 15.30 & 54.05 & 51.23 & 57.08 & +5.85 & 2.03 \\
Kimi-VL-A3B~\citep{team2025kimi} & 28.04 & 31.39 & 11.13 & 34.14 & 31.39 & 40.70 & \PABest{+9.31} & 4.77 \\
Molmo2-8B~\citep{clark2026molmo2} & 21.44 & 40.88 & 11.71 & 33.79 & 40.88 & 49.19 & +8.31 & 9.08 \\
\bottomrule
\end{tabular*}
\par\smallskip
\begin{minipage}{\linewidth}
\footnotesize\raggedright
Scores are percentages. $SolveAcc$ is mean normalized credit (Appendix~\ref{app:solve_scoring}).
$GAcc=GAcc_{\mathcal G}^{\mathrm{base}}$, $\Delta G$ is in percentage points.
Release pools: $\mathcal G$, 986 parents/3,341 probes, $\mathcal L$, 311/412, solving, 986 parents.
Base and +GT use identical probes and weights within each model:
304 parents/397 probes for each of GPT-6 Astra and Gemini-3.8-flash,
and 311/412 for each remaining model (Appendix~\ref{app:paired_support}).
Blue bold marks the column maximum across the ten models.
Changes are rounded independently from the displayed accuracies.
\end{minipage}
\endgroup
\end{table*}

\section{Experiments and Diagnostic Analysis}
\label{sec:experiments}

\subsection{Experimental Setup and Main Results}
\label{sec:experimental_setup}

We evaluated ten models: Qwen3.5-4B, -9B and -27B~\citep{qwen35blog},
InternVL3.5-8B~\citep{wang2025internvl3},
GPT-6 Astra~\citep{openai2026gpt6astra},
Gemini-3.8-flash~\citep{googledeepmind2026gemini38flash},
Gemma4-26B-A4B~\citep{team2026gemma}, GLM46V-flash~\citep{vteam2025glm45vglm41vthinkingversatilemultimodal}, Kimi-VL-A3B~\citep{team2025kimi} and Molmo2-8B~\citep{clark2026molmo2}.
Evaluation pools combine both release partitions.
Table~\ref{tab:main_results} reports local metrics on $\mathcal L$,
$\mathcal G$, and the model-specific paired sets $\mathcal P_m$,
alongside independent solving on the 986-problem release.
$SolveAcc$ averages normalized problem credit: binary credit for
choice and fill-in items, and expert-assigned partial credit for
short-answer and calculation items (Appendix~\ref{app:solve_scoring}).
Local metrics use balanced weights, solving weights parents equally.
Appendices~\ref{app:paired_support} and~\ref{app:model-configurations}
report observed paired coverage and inference configurations,
respectively, for all ten models.

Higher aggregate accuracy does not ensure consistent recovery of content
and its physical role. Within Qwen3.5, scaling from 4B to 27B increases
$CAcc$ from 53.97\% to 84.14\% and $JAcc$ from 32.09\% to 68.12\%.
GPT-6 Astra~\citep{openai2026gpt6astra} has the highest $CAcc$ (89.44\%) and $JAcc$ (77.13\%),
whereas Gemini-3.8-flash~\citep{googledeepmind2026gemini38flash} has the highest overall $GAcc$ (90.71\%).
Grounding and solving also differ: Gemma4-26B-A4B~\citep{team2026gemma} obtains 69.27\%
$GAcc$ but only 0.23\% $SolveAcc$.
These metrics describe distinct outcomes. To determine whether
grounding fails after a correct reading, we next examine both outcomes
on the same local probes.

\subsection{Recognition--Grounding Decomposition and Local-Reading Controls}
\label{sec:rg_decomp}

\begin{figure*}[t]
    \centering
    \includegraphics[width=\linewidth]{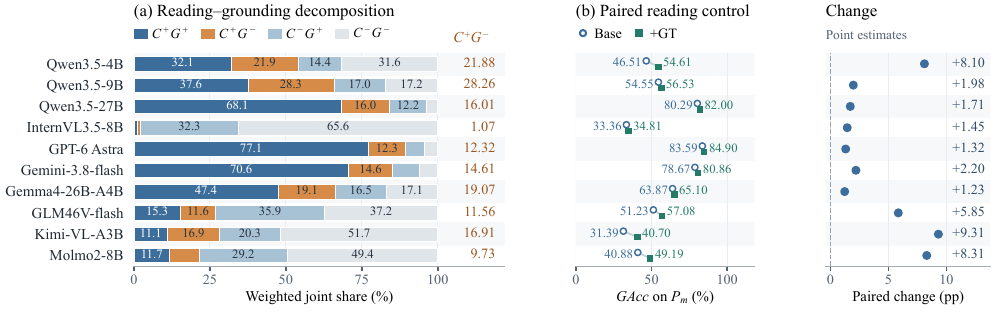}
    \caption{Recognition--grounding decomposition and paired controls for ten models.
        (a) Weighted joint outcomes on $\mathcal L$, reading-correct but grounding-wrong shares appear beside the bars.
        (b) Base/+GT grounding on model-specific observed sets $\mathcal P_m$ and paired changes in percentage points.
        Values are point estimates. +GT supplies local readings, not perfect perception.}
    \label{fig:decomposition}
\end{figure*}

Correctly reading local content does not guarantee correct grounding
of its physical role. Figure~\ref{fig:decomposition}(a) partitions
$\mathcal L$ into four recognition--grounding outcomes.
The weighted share of reading-correct but grounding-wrong events
($C^{+}G^{-}$) ranges from 1.07\% to 28.26\% across the ten models.
The shares are 19.07\% for Gemma4-26B-A4B~\citep{team2026gemma},
11.56\% for GLM46V-flash~\citep{vteam2025glm45vglm41vthinkingversatilemultimodal}, 16.91\% for Kimi-VL-A3B~\citep{team2025kimi} and 9.73\% for
Molmo2-8B~\citep{clark2026molmo2}. The same discrepancy therefore occurs across all evaluated
model families.

An event share must be distinguished from a conditional error rate.
InternVL3.5-8B~\citep{wang2025internvl3} has only 1.07\% $C^{+}G^{-}$, but also only 2.12\%
$CAcc$, its small event share does not indicate reliable grounding.
Conditioning instead on correct recognition, grounding error rates
range from 13.77\% for GPT-6 Astra~\citep{openai2026gpt6astra} to 60.30\% for Kimi-VL-A3B~\citep{team2025kimi}
(Appendix~\ref{app:pad1_full_results}). Even the lowest rate leaves
substantial disagreement between correct readings and reported roles.

\begin{table*}[t]
\centering
\caption{Grounding validity and robustness: rule-based baselines, image removal, and candidate reordering.}
\label{tab:validity_controls}
\begingroup
\small
\setlength{\tabcolsep}{3.2pt}
\renewcommand{\arraystretch}{1}
\begin{tabular*}{\linewidth}{@{\extracolsep{\fill}}lrrrr@{}}
\toprule
\multicolumn{5}{@{}l@{}}{\textbf{Rule-based baselines:}
Random candidates 18.69\%, nearest region 76.82\%.}\\
\midrule
\rowcolor{PAHeader}
\textbf{Model} & \makecell{\textbf{Reference}\\$GAcc\,\uparrow$} & \makecell{\textbf{No image}\\$GAcc\,\uparrow$} & \makecell{\textbf{No image}\\$\Delta G$ (pp)} & \makecell{\textbf{Reordered}\\$GAcc\,\uparrow$} \\
\midrule
Qwen3.5-4B~\citep{qwen35blog} & 40.21 & 38.78 & -1.43 & 41.93 \\
Qwen3.5-9B~\citep{qwen35blog} & 40.52 & 43.53 & +3.01 & 43.02 \\
Qwen3.5-27B~\citep{qwen35blog} & 84.11 & 63.72 & -20.39 & 84.48 \\
InternVL3.5-8B~\citep{wang2025internvl3} & 23.52 & 49.87 & +26.35 & 7.24 \\
GPT-6 Astra~\citep{openai2026gpt6astra} & 81.94 & 75.57 & -6.37 & 84.27 \\
Gemini-3.8-flash~\citep{googledeepmind2026gemini38flash} & \PABest{90.71} & \PABest{76.90} & -13.81 & \PABest{90.82} \\
Gemma4-26B-A4B~\citep{team2026gemma} & 69.27 & 59.83 & -9.44 & 74.59 \\
GLM46V-flash~\citep{vteam2025glm45vglm41vthinkingversatilemultimodal} & 54.05 & 47.47 & -6.58 & 52.49 \\
Kimi-VL-A3B~\citep{team2025kimi} & 34.14 & 35.45 & +1.31 & 32.79 \\
Molmo2-8B~\citep{clark2026molmo2} & 33.79 & 30.43 & -3.36 & 27.69 \\
\bottomrule
\end{tabular*}
\par\smallskip
\begin{minipage}{\linewidth}
\footnotesize\raggedright
Accuracies are percentages, changes are in percentage points.
Reference is $GAcc$ (all) in Table~\ref{tab:main_results}, Reordered permutes candidates.
$\Delta G=GAcc_{\mathrm{no\ image}}-GAcc_{\mathrm{base}}$.
Random and nearest-region scores retain their original applicability pools,
same-support comparisons appear in Appendix~\ref{app:geometry_hard}.
Blue bold marks column accuracy maxima,
signed changes are descriptive, not significance tests.
\end{minipage}
\endgroup
\end{table*}

Reference readings improve aggregate grounding in all models,
with gains from 1.23 points for Gemma4-26B-A4B~\citep{team2026gemma}
to 9.31 for Kimi-VL-A3B~\citep{team2025kimi}
(Figure~\ref{fig:decomposition}(b)).
These gains do not imply uniform probe-level benefits.
Paired transitions for three Qwen3.5 models and
InternVL3.5-8B~\citep{wang2025internvl3}
(Appendix~\ref{app:paired_transitions}) show both repairs and harms.
For Qwen3.5-4B~\citep{qwen35blog}, +GT repairs 13.35\% and
harms 5.25\% of weighted probe mass, a net gain of 8.10 points
(95\% CI [3.88, 12.50]), the other three net-change intervals
include zero.

Reference readings leave grounding errors in every model.
The highest +GT accuracy is 84.90\% for GPT-6 Astra~\citep{openai2026gpt6astra}, while
Qwen3.5-27B~\citep{qwen35blog} and Gemini-3.8-flash~\citep{googledeepmind2026gemini38flash} retain error rates of 18.00\% and
19.14\%, respectively, on their observed $\mathcal P_m$.
Since +GT withholds the correct owner or role and leaves other
visual and spatial uncertainties unresolved, it neither guarantees
role assignment nor defines a grounding upper bound.
The paired changes do not isolate a pure perception effect, and
residual errors cannot be attributed exclusively to physical reasoning.

\subsection{Validity: Visual Evidence, Geometry, and Interface Robustness}
\label{sec:validity}

Table~\ref{tab:validity_controls} examines sensitivity to visual input,
simple geometric regularities, and candidate presentation.
Each control addresses a distinct explanation for the grounding results.

\begin{figure*}[t]
    \centering
    \includegraphics[width=\linewidth]{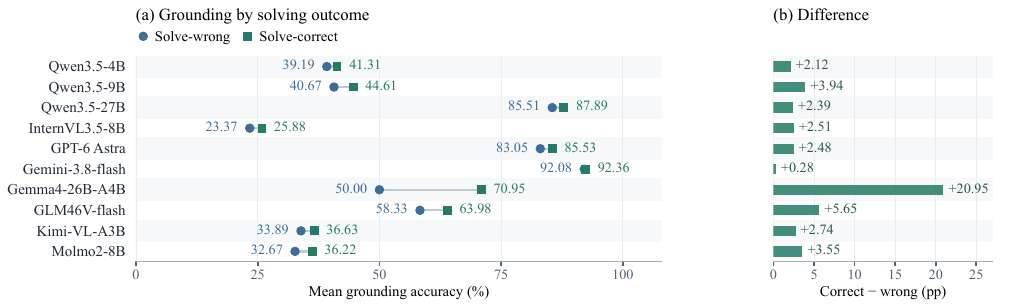}
    \caption{Local grounding by binary final-answer correctness for ten models, distinct from partial-credit $SolveAcc$.
        (a) Parent-level means for solve-wrong and solve-correct groups.
        (b) Correct-minus-wrong differences in percentage points.
        Values are point estimates from independent solving and probing runs, group membership is model-specific.}
    \label{fig:association}
\end{figure*}

Removing the image lowers $GAcc$ for seven models and raises it for
three. Qwen3.5-27B~\citep{qwen35blog}, GPT-6 Astra~\citep{openai2026gpt6astra}, and Gemini-3.8-flash~\citep{googledeepmind2026gemini38flash} lose 20.39,
6.37, and 13.81 percentage points, respectively. Gemma4-26B-A4B~\citep{team2026gemma},
GLM46V-flash~\citep{vteam2025glm45vglm41vthinkingversatilemultimodal}, and Molmo2-8B~\citep{clark2026molmo2} lose 9.44, 6.58, and 3.36 points.
Conversely, Qwen3.5-9B~\citep{qwen35blog}, InternVL3.5-8B~\citep{wang2025internvl3}, and Kimi-VL-A3B~\citep{team2025kimi} gain 3.01,
26.35, and 1.31 points. Visual-input sensitivity is therefore
model-dependent. Removing the image can also alter evidence anchors
and target identifiability, this control does not isolate the causal
contribution of visual reasoning.

The spatial baseline requires a comparison on identical probes and weights.
On the fixed applicable subset (665 parents, 1,639 probes), Qwen3.5-27B~\citep{qwen35blog}
achieves 86.35\% against 76.82\% for nearest-region, a 9.53-point
margin (95\% CI [6.10, 13.06]). On the same probes, the rule exceeds
Qwen3.5-4B, Qwen3.5-9B, and InternVL3.5-8B by 27.79, 25.04, and
46.52 points, respectively. We define \textit{H-hard} as probes
on which the frozen nearest-region rule is incorrect, independently
of model predictions (Appendix~\ref{app:geometry_hard}).
On this subset (271 parents, 394 probes), Qwen3.5-27B~\citep{qwen35blog} achieves 73.22\%
grounding accuracy against 18.76\% for random selection on the same probes.
On its joint-reading intersection (69 parents, 74 probes), the model's
$C^{+}G^{-}$ share remains 29.71\% [20.14, 39.84], with a 34.75\%
[23.77, 46.55] grounding error rate conditional on correct reading.
The tested proximity rule does not fully explain this model's
performance, and reading--grounding disagreement persists where
the rule fails. These results do not exclude other shortcuts.

Candidate reordering shifts aggregate $GAcc$ by at most 2.50 points
for seven of ten models, versus $-16.28$ for
InternVL3.5-8B~\citep{wang2025internvl3}, $+5.32$ for
Gemma4-26B-A4B~\citep{team2026gemma}, and $-6.10$ for
Molmo2-8B~\citep{clark2026molmo2}. Small aggregate shifts
do not establish permutation invariance: item-level changes
may cancel, and untested orders may yield different scores.

Together, these controls distinguish image-input sensitivity,
performance beyond the tested proximity rule, and sensitivity to
the tested candidate reordering. They motivate reporting geometry-defined
hard-subset results alongside aggregate grounding, without identifying
internal strategies or establishing a shortcut-free benchmark.

\subsection{Association with Original-Problem Solving}
\label{sec:association}

We compare parent-level grounding by binary final-answer correctness,
a diagnostic distinct from partial-credit $SolveAcc$.
Solving and probing use independent contexts, and the two outcome
groups are formed separately for each model.
Appendix~\ref{app:problem_solving} details the matching and aggregation.

All ten models show higher grounding point estimates on correctly
solved problems, but the gaps vary substantially.
Gemma4-26B-A4B~\citep{team2026gemma} shows a 20.95-point gap,
versus 0.28--5.65 points for the other nine models
(Figure~\ref{fig:association}(b)).
This consistent direction indicates a descriptive association
with model-dependent magnitude, but point estimates alone
establish neither statistical significance nor causality.

Correct final answers do not certify accurate local grounding.
On correctly solved problems, Qwen3.5-27B~\citep{qwen35blog}
and GPT-6 Astra~\citep{openai2026gpt6astra} show local grounding
error rates of 12.11\% and 14.47\%, respectively.
Conversely, Gemini-3.8-flash~\citep{googledeepmind2026gemini38flash}
achieves 92.08\% grounding accuracy on incorrectly solved problems
(Figure~\ref{fig:association}(a)).
High local grounding accuracy likewise cannot certify correct solutions.
PhysAlign complements solution accuracy by testing reported
evidence--role correspondences unresolved by final answers.
Our findings concern observable behavior: model-specific grouping
and independent contexts neither identify which local evidence
informed a solution nor establish that correcting a grounding error
would change the final answer.
Paired reading controls assess changes in reported grounding under
supplied local content with the target role withheld, not interventions
on the solver's actual premise use.

\noindent\textbf{Physical consequence of a role error.}
A cylinder of mass $m$ descends by height $h$ on a wedge of mass $M$.
The wedge moves horizontally, so only the cylinder loses gravitational
potential energy. Under both Base and +GT, Qwen3.5-9B correctly reads
$M$ but assigns it to the cylinder. We use this observed response to
compare two substitutions in the same energy relation. Using the
model's assignment for the descending mass gives $Mgh$. Using the
reference assignment gives $mgh$
(Appendix~\ref{app:casestudy}, Figure~\ref{fig:case-controlled-role-substitution}).

\FloatBarrier

\section{Conclusion}

PhysAlign complements solution accuracy with anchored probes that
separate content recognition from physical-role grounding.
Across ten models, correct recognition does not guarantee grounding,
reference readings resolve only some grounding errors, and correct
final answers can coexist with incorrect reported premises.
These findings motivate models that verify evidence--role
correspondences through query-conditioned local checks and are trained
with hard negatives that preserve content while swapping entities or roles.
Our findings concern explicit reports under fixed anchors and candidates
in static diagrams, not complete scene understanding, internal model
states, or premises used in a particular solution.
Domain coverage is uneven, and whether such targeted training improves
grounding and problem solving or transfers to dynamic settings
remains to be tested.
\FloatBarrier

{
  \small

}

\appendix
\phantomsection
\addcontentsline{toc}{section}{Supplementary Material}
\section*{Supplementary Material}
\section{Dataset Construction, Annotation, and Quality Assurance}
\label{app:dataset}

This appendix supplements the construction protocol in
Section~\ref{sec:probe} and the dataset overview in
Section~\ref{sec:quality_dataset}. Source coverage and split statistics
are reported in Table~\ref{tab:dataset_statistics} in the main text.
Here we provide provenance records, annotation details, review
procedures, release accounting, and a post-construction quality audit.
Unless an individual split or candidate pool is specified, counts refer
to the full retained release, combining development and test.
Evaluation pools are drawn from this release, observed paired coverage
is distinguished from release eligibility in
Appendix~\ref{app:paired_support}. Split-specific counts describe the
release inventory, not separate performance evaluations.
Section~\ref{app:quality_audit} reports an audit of the retained
inventory and the disposition of flagged probes in a subsequent
release candidate.

\subsection{Source Problems and Provenance}
\label{app:problem_data}

\paragraph{Source records.}
Table~\ref{tab:dataset_statistics} lists the six source datasets and their
retained parent counts, rather than the sizes of the upstream datasets.
For OlympiadBench~\citep{he2024olympiadbench}, we use its English physics
competition subset.
Each source record preserves the original image, statement, question,
answer options when present, reference answer or solution, and provenance
identifiers.
PhysAlign augments these records with evidence-linked observational
annotations, local probes, candidate mappings, recognition packets,
and scoring records.
The original physics problems are reused rather than newly authored.

\paragraph{Deduplication and split propagation.}
Multiple probes can originate from the same parent problem and therefore
do not represent independent source problems.
The merge procedure deduplicates exactly identical public inputs and
propagates split assignments and isolation decisions through
original-image hashes and existing recorded groups.
The current merge introduced no additional exclusions.
All retained probes from a parent remain in the same split, with
membership recorded in \texttt{private/membership.json}.
These checks address exact-input duplication and recorded grouping,
they do not establish the absence of semantic near-duplicates or prior
model exposure to the original problems.

\subsection{Observational Annotation and Probe Compilation}
\label{app:observational_schema}

\paragraph{Pass-level outputs.}
The observational stage in Figure~\ref{fig:construction} comprises four
answer-blind passes.
Pass~1 records visible primitives, text, and locations.
Pass~2 links them to physical entities.
Pass~3 records textual mentions, explicit quantities, constraints, and
query targets. And Pass~4 consolidates directly supported observations
and evidence links into $G_{\mathrm{obs}}$.
Correct answers and reference solutions are excluded from these four passes.

Pass~5 aligns atomic steps of an existing reference solution with the
observations they use.
It is reserved for analysis: it neither re-solves the problem nor supplies
probe targets or solution-derived corrections to $G_{\mathrm{obs}}$.

\paragraph{Exported fields and mappings.}
The interfaces in Section~\ref{sec:task} export \texttt{referent} for
\texttt{T02-single}, \texttt{read} and \texttt{owner} for
\texttt{T03-image}, and \texttt{owner} for \texttt{T03-text}.
Each instance associates an evidence anchor ($R$), neutral candidate
entities ($E$), and an original or localized view ($V$).
Targets are compiled from reviewed observational fields, while
alias-to-entity mappings remain on the scoring side.

Templates preserve the physical meaning of their source fields.
In particular, a quantity's \texttt{owner\_id} identifies the entity to
which that quantity belongs, the owner may be an object, force, field,
or another physical entity.
Identifying the force entity whose magnitude is shown, for example,
differs from identifying the object on which that force acts.
The compiler does not substitute one relation for the other.
Appendix~\ref{app:probe_scoring} specifies the input/output interfaces
and +GT information rules.

\subsection{Quality Control and Admission Rules}
\label{app:review}

\paragraph{Verification and correction.}
Quality control checks both the source annotations and the exported probes.
Source verification compares transcription, localization, entity
correspondence, quantity ownership, and evidence support against the
original problem.
Exported-probe review examines the complete public question, evidence
anchor, candidate set, and scoring target.
Structural checks validate references, dependencies, mappings, and
localization. Semantic correctness, answer uniqueness, and candidate
distinguishability require review against the original text and images.

Human verification follows the blind-first procedure in
Section~\ref{sec:probe}.
Two coauthors independently answer the public probe before consulting
the private target and source annotations.
Direct approval requires agreement between their independent answers
and successful quality checks, disagreements proceed to a third reviewer
for adjudication.
Review also checks candidate completeness, transcription, localization,
and information leakage.
Corrections require regeneration and rechecking, while shared-source or
rule-level issues require examination of related instances.
The construction-stage procedures above are distinct from the
post-construction audit of retained probes in
Section~\ref{app:quality_audit}.

\paragraph{Revision of further-review candidates.}
We used GPT-6 Astra~\citep{openai2026gpt6astra} to revise the 1,089 candidates marked as requiring
further review at initial screening.
The first stage checked probe content against the original text and
images and corrected transcription and reading-target errors where
applicable.
The second stage compared evidence regions, candidate entities, and
target relations with the original images to determine the required
localization and semantic corrections.
The revised probes then underwent item-by-item human verification using
the same blind-first procedure.
These decisions apply to the revised versions, the original screening
states remain part of the initial candidate accounting.
Section~\ref{app:exclusion_analysis} reports the final review outcomes
and distinguishes the supplementary candidates from the retained benchmark.

\paragraph{Admission and audit interpretation.}
The recorded construction-stage test-parent admission condition is
\[
n_i^{\mathrm{rej}} \leq 2,
\]
where $n_i^{\mathrm{rej}}$ counts generated candidate probes for parent $i$
recorded as rejected.
This condition governs parent eligibility, it neither admits rejected
probes nor overrides subsequent risk isolation.
Candidates with unresolved screening states or incomplete calls were
excluded from the reported benchmark.
Subsequent revisions and their review outcomes are accounted for separately.

Table~\ref{tab:app-screening-retention}(a) records model-screening states,
whereas panel~(c) reports final human-review outcomes for revised probes.
Model approval does not establish agreement between human reviewers,
and a further-review state does not itself indicate human disagreement.
Neither the screening counts nor final post-revision approvals are used
to infer pre-adjudication agreement or a human-performance ceiling.

\paragraph{Input--scoring separation.}
Model-visible files contain base and +GT inputs together with the
images available for inference.
Reference targets, source and candidate mappings, parent membership,
source snapshots, and review records are stored separately for scoring
and audit.

\subsection{Candidate Screening and Release Retention}
\label{app:exclusion_analysis}

Table~\ref{tab:app-screening-retention}(a) accounts for 5,594 generated
probes from 1,084 candidate parents.
Initial model screening yielded 3,370 approvals, 1,128 rejections,
1,089 candidates requiring further review, and seven incomplete or failed
screening calls.
These four states are mutually exclusive.
Risk isolation subsequently excluded 29 model-approved candidates,
leaving 3,341 retained probes from 986 parents, as shown in panel~(b).
The retained probes constitute 59.72\% of generated candidates.
The other 2,253 candidates were not admitted to the reported benchmark
at that stage: 1,128 model rejections, 1,089 further-review cases,
seven incomplete calls, and 29 risk-isolated approvals.
These initial states do not imply that every excluded candidate was
semantically incorrect, subsequent revisions are reported separately.

\begin{table}[t]
    \centering
    \small
    \setlength{\tabcolsep}{4pt}
    \renewcommand{\arraystretch}{1.15}

    \caption{
        \textbf{Candidate-probe screening, release admission, and human re-review.}
        Panel (a) records the initial model-screening states of 5,594
        generated probes, panel (b) reports admission within the
        model-approved subset. Panel (c) reports final human-review
        outcomes for revised versions of the 1,089 candidates initially
        requiring further review.
    }
    \label{tab:app-screening-retention}

    \begin{tabularx}{\linewidth}{
        >{\raggedright\arraybackslash}Xr
    }
        \toprule

        \rowcolor{black!5}
        \textbf{Recorded status or outcome}
        & \textbf{Probes}
        \\

        \midrule

        \rowcolor{black!5}
        \multicolumn{2}{@{}l}{
            \textbf{(a) Initial model-screening states}
        }
        \\

        Approved by model screening
        & 3,370 \\

        Rejected by model screening
        & 1,128 \\

        Further review required at initial screening
        & 1,089 \\

        Incomplete / call error
        & 7 \\

        \midrule

        \textbf{Total generated candidates}
        & \textbf{5,594} \\

        \addlinespace[4pt]

        \rowcolor{black!5}
        \multicolumn{2}{@{}l}{
            \textbf{(b) Admission within model-approved candidates}
        }
        \\

        Withheld through risk isolation
        & 29 \\

        \textbf{Retained probes}
        & \textbf{3,341} \\

        \addlinespace[4pt]

        \rowcolor{black!5}
        \multicolumn{2}{@{}l}{
            \textbf{(c) Human re-review of revised candidates}
        }
        \\

        Revised probes reviewed by humans
        & 1,089 \\

        Approved after review, retained in supplementary pool
        & 1,089 \\

        Rejected or unresolved after review
        & 0 \\

        Added to the reported benchmark
        & 0 \\

        \bottomrule
    \end{tabularx}

    \vspace{2pt}

    \begin{minipage}{\linewidth}
        \footnotesize
        The states in (a) sum to 5,594. The 29 isolated candidates are
        included in the 3,370 approvals, so (b) sums to
        $29+3{,}341=3{,}370$, panels must not be added together.
        Panel (c) follows revised versions of the 1,089 further-review
        candidates in (a). Its coverage and outcome rows describe the
        same probes and must not be summed. These revised probes form a
        separate supplementary candidate pool and were not used in the
        reported evaluation. Human outcomes are final post-revision
        decisions, not first-pass acceptance or inter-reviewer agreement.
        Incomplete calls concern construction-stage screening.
    \end{minipage}

\end{table}

Among the 986 retained parents, 184 retain their complete generated probe
set, whereas 802 retain only an approved subset.
Here, \emph{complete} means that all generated candidates for the parent
are retained, not that every physical fact in the problem is annotated.
Partial retention means that only probes satisfying admission requirements
enter the benchmark, candidates rejected or unresolved at screening
were not admitted in those versions.
The remaining $1{,}084-986=98$ candidate parents contribute no retained
probe. The available accounting does not attribute this parent-level
difference to a single exclusion cause.

\paragraph{Human re-review and supplementary candidates.}
Panel~(c) reports the outcomes of the revision procedure in
Section~\ref{app:review}.
All 1,089 revised probes underwent item-by-item human review and received
final approval, none remained rejected or unresolved.
These are final outcomes after revision, not first-pass acceptance rates
or measures of inter-reviewer agreement.
The approved revisions form a separate supplementary candidate pool
and were not used in the reported model evaluation.
Dataset counts, evaluation denominators, and experimental results
therefore refer to the retained 3,341 probes from 986 parents.
They characterize these retained local tasks rather than the full
unfiltered candidate pool.

\subsection{Post-construction Quality Audit}
\label{app:quality_audit}

\paragraph{Sampling protocol.}
The sampling frame for the audit in Section~\ref{sec:probe} comprised
the retained probes. Section~\ref{app:exclusion_analysis} describes
the separate review of 1,089 supplementary candidates.
Table~\ref{tab:app-quality-audit}(a) gives the allocation by audit
source group. All 201 probes in the three smallest groups were
inspected. The other 2,199 slots were allocated approximately in
proportion to the sizes of the three largest groups. Within each of
these groups, a fixed-seed hash ordering of \texttt{logical\_probe\_id}
determined the sample without replacement. Selection was independent
of model scores, problem difficulty, current gold answers, and
previous review outcomes.

\begin{table*}[t]
\centering
\caption{%
Post-construction quality audit of retained probes.
(a) Sampling allocation. (b) Review outcomes within the audited set.
}
\label{tab:app-quality-audit}
\begingroup
\small
\setlength{\tabcolsep}{4pt}
\renewcommand{\arraystretch}{1.16}

\begin{minipage}[t]{0.49\textwidth}
\vspace{0pt}
\textbf{(a) Sampling allocation}\par\vspace{4pt}
\begin{tabularx}{\linewidth}{@{}>{\raggedright\arraybackslash}Xrrr@{}}
\toprule
\rowcolor{black!5}
\textbf{Audit source group}\textsuperscript{a}
& \shortstack[r]{\textbf{Total}\\\textbf{probes}}
& \textbf{Audited}
& \shortstack[r]{\textbf{Coverage}\\\textbf{(\%)}} \\
\midrule
Local          & 1,679 & 1,176 & 70.04 \\
PhysElite      &   792 &   554 & 69.95 \\
OlympiadBench  &   669 &   469 & 70.10 \\
\addlinespace[2pt]
Expansion-v3   &   123 &   123 & 100.00 \\
Phyx-OE        &    40 &    40 & 100.00 \\
Gaokao         &    38 &    38 & 100.00 \\
\midrule
\rowcolor{black!8}
\textbf{Total}
& \textbf{3,341} & \textbf{2,400} & \textbf{71.83} \\
\bottomrule
\end{tabularx}
\end{minipage}\hfill
\begin{minipage}[t]{0.485\textwidth}
\vspace{0pt}
\textbf{(b) Audit disposition}\par\vspace{4pt}
\begin{tabularx}{\linewidth}{@{}>{\raggedright\arraybackslash}Xrr@{}}
\toprule
\rowcolor{black!5}
\textbf{Review outcome}\textsuperscript{b}
& \textbf{Probes} & \shortstack[r]{\textbf{Share}\\\textbf{(\%)}} \\
\midrule
Passed first review without revision
& 2,332 & \textbf{97.17} \\
Passed after correction and reinspection
& 66 & 2.75 \\
Unresolved after reinspection, excluded
& 2 & 0.08 \\
\midrule
\rowcolor{black!8}
\textbf{Final accepted within audited set}\textsuperscript{c}
& \textbf{2,398} & \textbf{99.92} \\
\bottomrule
\end{tabularx}
\end{minipage}
\par\vspace{4pt}
\begin{minipage}{\linewidth}
\footnotesize\raggedright
\textsuperscript{a} Labels follow the audit report. Counts refer to probes, not the
parent-level source counts in Table~\ref{tab:dataset_statistics}.
\par
\textsuperscript{b} The first three rows partition the 2,400 audited probes.
All shares in (b) use 2,400 as the denominator.
\par
\textsuperscript{c} Final acceptance combines the first two outcome rows, it is not
an additional category or an estimate of full-dataset accuracy.
\par
\end{minipage}
\endgroup
\end{table*}


\paragraph{Correction and reinspection.}
Flagged transcription issues included missing symbols, subscripts,
superscripts, and units. Other flags concerned evidence locations
or inconsistencies among owners, candidates, and private answer mappings.
GPT-6 Astra~\citep{openai2026gpt6astra}, run at high reasoning effort,
first corrected local readings and displaced reading locators, then
checked evidence regions, candidate entities, ownership, and gold
answers against the source images. Reinspection applied the original
criteria for transcription, localization, ownership, and answer mapping.
Table~\ref{tab:app-quality-audit}(b) records the dispositions.
The audit trail for each flagged probe includes its identifier, source
image, original annotation, issue description, correction history,
and reinspection decision.

\paragraph{Interpretation and release scope.}
First-pass acceptance measures the share accepted before audit-driven
correction. Final acceptance also includes probes that passed after
correction. Both rates use the 2,400 audited probes as the denominator.
Sampling fractions vary by source group, and 941 probes were outside
the audit. The pooled rates describe the inspected sample and do not
measure inter-reviewer agreement, human performance, or model accuracy.
The exclusion of unresolved probes applies to the subsequent release
candidate. The evaluation in Section~\ref{sec:experiments} uses the
reported 3,341-probe inventory. The audit does not quantify the effect
of the corrections on those model scores.

\subsection{Composition, Problem Content, and Evaluation Coverage}
\label{app:dataset_statistics}

\begin{table*}[t]
\centering
\caption{
Estimated domain composition and parent-level paired-input coverage of PhysAlign.
Counts refer to parent problems, domain assignments are estimated, and paired-input
coverage is measured within each group.
}
\label{tab:domain_composition_paired}

\small
\setlength{\tabcolsep}{5pt}
\renewcommand{\arraystretch}{1.16}

\begin{tabularx}{\linewidth}{
    >{\raggedright\arraybackslash}X
    rrrrr
}
\toprule

& \multicolumn{2}{c}{\textbf{All parents}}
& \multicolumn{3}{c}{\textbf{Paired-input coverage}} \\

\cmidrule(lr){2-3}
\cmidrule(lr){4-6}

\textbf{Domain group}
& \textbf{Count}
& \textbf{Share (\%)}
& \textbf{Paired}
& \textbf{Unpaired}
& \textbf{Rate (\%)} \\

\midrule


\rowcolor{black!6}
Mechanics$^{a}$
& 478 & 48.48 & 208 & 270 & 43.51 \\

\rowcolor{black!6}
Electromagnetism$^{a}$
& 267 & 27.08 & 112 & 155 & 41.95 \\

Optics
& 50 & 5.07 & 15 & 35 & 30.00 \\

Thermal physics / thermodynamics
& 48 & 4.87 & 10 & 38 & 20.83 \\

Waves and acoustics
& 35 & 3.55 & 9 & 26 & 25.71 \\

Modern, atomic and nuclear physics
& 19 & 1.93 & 4 & 15 & 21.05 \\

Astronomy and astrophysics
& 11 & 1.12 & 1 & 10 & 9.09 \\


\midrule
\multicolumn{6}{l}{
    \textit{Residual classification groups}$^{b}$
} \\

\addlinespace[1pt]

Cross-domain / uncertain
& 60 & 6.09 & 21 & 39 & 35.00 \\

Unclassified
& 18 & 1.83 & 5 & 13 & 27.78 \\


\midrule

\rowcolor{black!9}
\textbf{Full release}
& \textbf{986}
& \textbf{100.00}
& \textbf{385}
& \textbf{601}
& \textbf{39.05} \\

\bottomrule
\end{tabularx}

\vspace{4pt}

\begin{minipage}{\linewidth}
\footnotesize

\noindent\textit{Definitions.}
For group $d$, let $N_d$ denote its parent count and $H_d$
the number of parents with at least one retained paired +GT
variant.
\textit{Share} $=100N_d/986$,
\textit{Rate} $=100H_d/N_d$,
\textit{Unpaired} $=N_d-H_d$.
Percentages are rounded to two decimals and may not sum to 100.

\par\noindent
$^{a}$ Shaded rows highlight sample concentration:
mechanics and electromagnetism together account for approximately
75.6\% of all parents and 83.1\% of parents with paired +GT
variants.

\par\noindent
$^{b}$ These are residual classification labels rather than
additional physics domains. All domain-level counts and derived
percentages use estimated assignments.

\end{minipage}
\end{table*}

\begin{figure*}[!t]
    \centering
    \includegraphics[width=\textwidth]{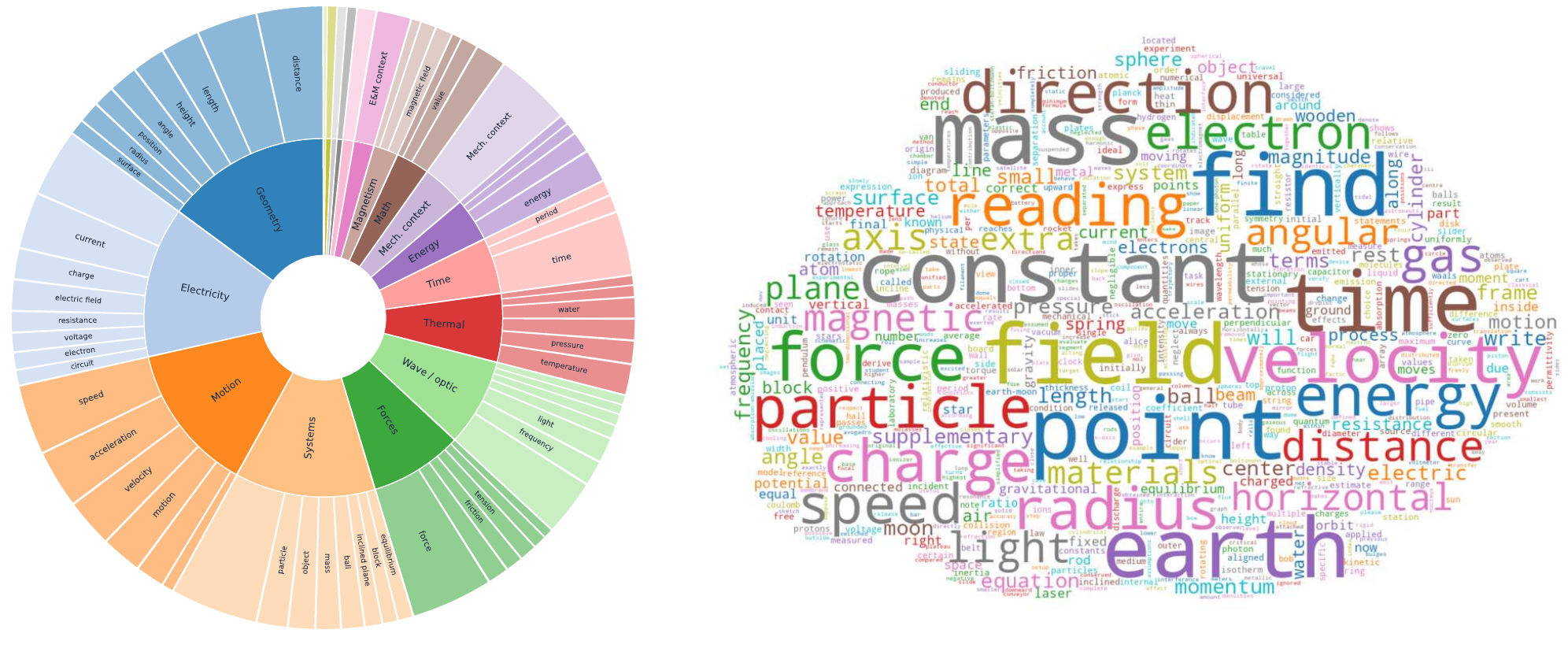}
    \caption{
        \textbf{Problem-content profile of the retained PhysAlign source
        problems.}
        The left panel hierarchically groups recurring physics-related
        vocabulary into broad concept families, while the right panel
        visualizes prominent terms appearing in the problem text.
        This lexical view complements the parent-level domain statistics
        in Table~\ref{tab:domain_composition_paired}.
        It is not used for domain assignment, probe construction,
        eligibility filtering, or benchmark scoring.
    }
    \label{fig:problem-content}
\end{figure*}

\begin{table*}[t]
\centering
\caption{
Probe density, paired-input coverage, and release statistics of PhysAlign.
Panel (a) summarizes split-level statistics, panel (b) reports
full-release quantities and counting units.
}
\label{tab:probe_density_and_coverage}

\small
\setlength{\tabcolsep}{5pt}
\renewcommand{\arraystretch}{1.10}

\begin{tabularx}{\linewidth}{@{}
    >{\raggedright\arraybackslash}X
    >{\raggedleft\arraybackslash}p{0.22\linewidth}
    >{\raggedleft\arraybackslash}p{0.27\linewidth}
    @{}}
\toprule
\multicolumn{3}{@{}l}{
    \textbf{(a) Split-level density and paired-probe coverage}
} \\
\addlinespace[3pt]
\textbf{Split}
    & \textbf{Probes / parent}
    & \textbf{Paired coverage (\%)}$^{a}$ \\
\midrule
Development
    & 2.63
    & 43.27 \\
Test
    & 3.45
    & 14.78 \\
\midrule
\textbf{Full release}
    & \textbf{3.39}
    & \textbf{16.55} \\
\bottomrule
\end{tabularx}

\vspace{8pt}

\begin{tabularx}{\linewidth}{@{}
    >{\raggedright\arraybackslash}X
    >{\raggedleft\arraybackslash}p{0.12\linewidth}
    >{\raggedright\arraybackslash}p{0.31\linewidth}
    @{}}
\multicolumn{3}{@{}l}{
    \textbf{(b) Full-release statistics and counting units}
} \\
\addlinespace[3pt]
\textbf{Statistic}
    & \textbf{Value}
    & \textbf{Denominator / unit} \\
\midrule
Parents with paired +GT inputs
    & 39.05\%
    & 385/986 parents \\
Paired +GT variants per paired parent$^{b}$
    & 1.44
    & 553/385 \\
Localized probes per parent: median, range
    & 3, 1--6
    & Parent-level summary \\
Localized probes without a paired +GT variant
    & 2,788
    & 3,341 $-$ 553 probes \\
Packaged image files$^{c}$
    & 18,640
    & File-level count \\
\bottomrule
\end{tabularx}

\vspace{4pt}

\begin{minipage}{\linewidth}
\footnotesize

\noindent\textit{Definitions.}
For split $s$, $N_s$ denotes the number of parents,
$\mathcal{G}_s$ the localized-probe pool, and
$\mathcal{P}_s \subseteq \mathcal{G}_s$ the paired subset,
density is $|\mathcal{G}_s|/N_s$.
Paired +GT variants are not additional unique probes.

\par\noindent
$^{a}$ Coverage is
$100|\mathcal{P}_s|/|\mathcal{G}_s|$:
$90/208$ (development),
$463/3{,}133$ (test), and
$553/3{,}341$ (overall).

\par\noindent
$^{b}$ The denominator comprises the 385 parents with
paired +GT variants, not all 986, reported Base/+GT
scores use observed subsets $\mathcal{P}_m$
(Appendix~\ref{app:paired_support}).

\par\noindent
$^{c}$ The 18,640 files are not 18,640 unique source
diagrams or independent physics problems.

\end{minipage}
\end{table*}

\paragraph{Counting units and paired coverage.}
The following statistics describe the retained benchmark, the
supplementary candidate pool is excluded.
We use the evaluation sets defined in Section~\ref{sec:scoring}.
The retained dataset contains $|\mathcal G|=3{,}341$ unique grounding
probes, each with a base input.
A subset $\mathcal P\subseteq\mathcal G$ with $|\mathcal P|=553$ also
has a paired +GT input variant.

The 553 +GT instances are paired input variants rather than additional
independent probes.
Similarly, the \texttt{read} and \texttt{owner} outputs of a
\texttt{T03-image} instance are two fields of one probe rather than
two separate probes.
The remaining $3{,}341-553=2{,}788$ localized probes are valid retained
tasks without a paired +GT variant.They should not be confused with
the 2,253 original candidates excluded from the reported release.

There are 385 parent problems with at least one retained paired +GT
variant and 601 without one.
Parent-level coverage is $385/986\approx39.05\%$, whereas probe-level
paired coverage is $553/3{,}341\approx16.55\%$.
These ratios answer different questions and are neither model-accuracy
scores nor annotation-agreement measures.
The joint reading--grounding set $\mathcal L$ contains 412 probes
from 311 parents. Its reading-target eligibility differs from the
paired-input eligibility of $\mathcal P$ (553 probes from 385 parents),
these two evaluation populations must not be conflated.

\paragraph{Estimated domain composition.}
Table~\ref{tab:domain_composition_paired} reports the estimated parent-level
domain composition of the retained benchmark.
It distinguishes seven named physics-domain groups from the residual
classifications ``Cross-domain / uncertain'' and ``Unclassified.''
The assignments are descriptive estimates rather than a verified
fine-grained subject taxonomy.
Figure~\ref{fig:dataset-representation} combines the two residual groups
as ``Cross-disciplinary'' (78 parents), whereas
Table~\ref{tab:domain_composition_paired} reports them separately.

Mechanics and electromagnetism contribute 478 and 267 parents,
respectively, accounting for approximately 75.6\% of the release.
Their 208 and 112 parents with paired +GT variants account for
approximately 83.1\% of parents with paired variants.
Neither the full parent distribution nor its paired-input coverage
is therefore domain-balanced.
These are parent-level statistics and do not determine the domain
distribution of the 553 paired probes, since a parent may contribute
multiple +GT variants.



\paragraph{Problem-content vocabulary and concept coverage.}
The coarse domain labels summarize where the source problems belong,
but do not identify the physical quantities, objects, relations, and
geometric concepts within them.
Figure~\ref{fig:problem-content} complements the domain inventory with
a lexical view of the retained source problems.
The left panel organizes recurring problem vocabulary into concept
families, the right panel shows prominent physics-related terms in
the problem text.
This analysis is descriptive: lexical frequency is not used to assign
domain labels, construct candidates, determine probe eligibility,
or compute evaluation scores.

\paragraph{Probe density and split-specific coverage.}
Table~\ref{tab:probe_density_and_coverage} reports probe density and paired
coverage for the development and test splits.
For split $s$, localized-probe density is $|\mathcal G_s|/N_s$ and
paired coverage is $100|\mathcal P_s|/|\mathcal G_s|$, where $N_s$ is
the number of retained parent problems.
These are inventory summaries rather than performance metrics.

Across the retained release, the mean localized-probe count per parent
is $3{,}341/986\approx3.39$, with a median of 3 and a range of 1--6.
Among the 385 parents with paired +GT variants, the mean paired-variant
count is $553/385\approx1.44$, this denominator excludes parents
without paired variants.

Using the split counts in Table~\ref{tab:dataset_statistics}, development
and test have localized-probe densities of 2.63 and 3.45 probes per
parent and paired-probe coverage rates of 43.27\% and 14.78\%, respectively.

The release contains 553 paired-eligible probes from 385 parents in
$\mathcal P$. Reported Base/+GT comparisons use the observed
\texttt{T03-image} subsets $\mathcal P_m\subseteq\mathcal P$, with
identical probes and weights across conditions within each model
(Appendix~\ref{app:paired_support}), not separate split-level evaluations.
Comparing +GT on $\mathcal P_m$ with Base on all of $\mathcal G$ would
confound the input intervention with a change in evaluation population.



Balanced performance aggregation and matched base/+GT comparisons
follow Section~\ref{sec:scoring}.

\FloatBarrier

\section{Probe Specifications and Scoring}
\label{app:probe_scoring}

This appendix explains how local tests are presented, how model outputs are converted into scores, and how results on different evaluation sets are aggregated. Data sources, annotation review, and admission statistics are provided in Appendix~\ref{app:dataset}. Prompt templates, model configurations, and execution records are provided in Appendix~\ref{app:prompts-and-implementation}. Local-probe scoring covers the \texttt{T02-single}, \texttt{T03-image},
and \texttt{T03-text} interfaces. Independent original-problem scoring
is specified in Appendix~\ref{app:solve_scoring}.

\subsection{Public Inputs, Evidence Anchors, and Candidates}
\label{app:public_inputs}

Each public probe comprises the original problem text and diagram,
a local question, an evidence anchor, candidates, and output requirements.
Image anchors identify explicit regions in a specified original image,
text anchors identify exact spans in the original text.
Localized views display these locations without modifying the original
problem content.
Candidates are presented through probe-specific neutral aliases linked
to their image regions or text spans.
Mappings from aliases to reference entities are retained on the scoring
side, models need not generate internal entity identifiers.

In the static-equilibrium example in Figure~\ref{fig:example}, anchor R1 marks the
symbol \(T\), while E2, E3, and E4 identify the block, rod, and supporting
cable, respectively.
The local question asks ``What is the symbol in R1, and to which entity
does it belong?''
The reference associates \(T\) with the supporting cable (E4), whose
tension it denotes.
The illustrated response instead reports \texttt{read}: \(T\) and
\texttt{owner}: E3 (rod).
The resulting scores are \(C_j=1\), \(G_j=0\), and \(J_j=0\):
content recognition is correct, but the physical ownership of the
recognized symbol is not preserved.
Entity identifiers are used here for exposition, actual responses use
the corresponding public candidate aliases.
The reference targets and scores shown in the figure are evaluation
annotations, not inputs to the illustrated probe.
When an eligible probe has a paired +GT variant, that variant may
additionally supply ``the symbol in R1 is \(T\)'', but must not reveal
its ownership, such as ``\(T\) denotes the tension in the supporting
cable.''

Candidate construction and display order are fixed before evaluation.
Single-target queries contain at least two valid candidates, and the
correct target must have a valid, distinguishable public representation.
Existing labels in the original image are retained, but answer-revealing
descriptions derived from private annotations, such as
``E4 is the cable associated with \(T\)'', are not added.
Unresolved entity correspondences, ambiguous references, and
indistinguishable candidates are not forced into single-target queries.
\texttt{T02-single} returns \texttt{referent},
\texttt{T03-image} returns \texttt{read} and \texttt{owner},
and \texttt{T03-text} returns \texttt{owner}.
Reusable prompts are provided in
Appendix~\ref{app:prompts-and-implementation}.

\subsection{Ground-Truth Local Content and Evaluation Pools}
\label{app:gt_content}

The Base condition (also labeled Raw in the figures) and the +GT
condition use the same original text and diagram, question, anchor,
candidates, and output requirements, and are run in independent contexts.
+GT only adds predefined ground-truth local recognition information.
The supplied content is selected by rules independent of the tested model and
its errors, and must neither directly give the tested role nor select only the
correct candidate's label or leak associations through field combinations.
Whether content may be supplied depends on the current query target,
a reviewed fact is not automatically admissible as +GT input. Reproducing content supplied under +GT is
not counted as new recognition ability.

Let \(\mathcal{G}\) denote all eligible grounding tests,
\(\mathcal{L} \subseteq \mathcal{G}\) the joint evaluation set with
independent reading targets, and \(\mathcal{P} \subseteq \mathcal{G}\) the
paired set with non-empty, valid, and permitted GT content. In the current
interfaces, \texttt{T03-image} items that satisfy reading eligibility can
enter \(\mathcal{L}\), \texttt{T02-single} and \texttt{T03-text} do not
receive reading scores merely because the original text contains numerical
values. The eligibility of \(\mathcal{L}\) and \(\mathcal{P}\) is determined
separately, and the size of the joint set cannot be inferred from the number
of paired instances. The absence of valid GT content does not necessarily
remove an otherwise valid grounding test from \(\mathcal{G}\), but that item
cannot be counted toward the coverage of the recognition-augmented comparison.

Evaluation pools are drawn from the full retained release, combining
development and test. The following counts describe release eligibility
and the independent-solving population: $N$ counts distinct parent
problems and $n$ counts unique probes.
\begin{center}
\footnotesize
\setlength{\tabcolsep}{2.5pt}
\renewcommand{\arraystretch}{1.12}
\begin{tabularx}{\linewidth}{@{}>{\raggedright\arraybackslash}p{0.38\linewidth}rr>{\raggedright\arraybackslash}X@{}}
\toprule
\rowcolor{black!5}
Release pool & $N$ & $n$ & Scope \\
\midrule
Overall grounding $\mathcal G$ & 986 & 3,341 & $GAcc$ \\
Joint reading--grounding $\mathcal L$ & 311 & 412
  & $CAcc$, $JAcc$,\newline $GAcc_{\mathcal L}^{\mathrm{base}}$ \\
Paired-eligible release $\mathcal P$ & 385 & 553
  & Valid +GT input variants \\
Original-problem solving & 986 & --- & $SolveAcc$ \\
\bottomrule
\end{tabularx}
\end{center}
The release sets $\mathcal L$ and $\mathcal P$ have distinct eligibility
criteria and are not equal. Reported paired scores instead use model
$m$'s observed subset $\mathcal P_m\subseteq\mathcal P$, with identical
Base/+GT probes and weights within that model.
Appendix~\ref{app:paired_support} defines this subset, reports coverage
for all ten models, and explains when its Base score equals that on $\mathcal L$.
The partition in Table~\ref{tab:dataset_statistics} is inventory
metadata, the results are not held-out test estimates.

\subsection{Deterministic Scoring, Aggregation, and Uncertainty}
\label{app:scoring_details}

\paragraph{Parsing and Normalization.}
For local probes, the model should return a single JSON object as declared by the task. The scorer accepts this object or a single code block wrapping only this object, and rejects duplicate keys and multiple objects that cannot be uniquely determined, extra fields do not participate in target scoring. Completed and parseable responses are scored independently by field: a missing \texttt{read} does not cause a valid and correct \texttt{owner} to be zeroed, and an invalid candidate does not erase an independently correct reading. The content normalization rule \(N_j\) is fixed before evaluation and only handles declared format equivalences, preserving differences in digits, signs, case, subscripts, superscripts, and units. For example, a transcription rule that explicitly allows spaces between a number and a unit may accept ``2kg'' and ``2 kg'', but does not accept ``2000 g'' in place of ``2 kg'' merely because of physical numerical equivalence. No other language model is used to repair answers or to adjust matching rules post hoc.

\paragraph{Item-level Events.}
Let \(\hat{c}_j\) and \(\hat{u}_j\) be the predicted reading and candidate
alias, \(c_j\) and \(r_j\) the reference content and role target, and
\(\phi_j\) the mapping from candidates to reference entities, and \(E_j\)
the set of public candidate aliases. For scorable fields, define
\begin{equation}
\begin{aligned}
C_j &= \mathbf{1}[N_j(\hat{c}_j) = N_j(c_j)], \\
G_j &= \mathbf{1}[\hat{u}_j \in E_j \land \phi_j(\hat{u}_j) = r_j], \\
J_j &= C_j G_j.
\end{aligned}
\end{equation}
In the following, \(C_j, G_j, J_j\) default to the base condition.
\(C_j\) and \(J_j\) apply only to \(\mathcal{L}\), the +GT condition
evaluates only \(G_j^{+\mathrm{GT}}\), and its C/J and the C/J of
grounding-only items are recorded as not applicable rather than zero.

\paragraph{Parent-problem and Type Balancing.}
For a frozen set \(\mathcal{S}\), let \(\mathcal{S}_{it}\) be the probes of parent problem \(i\) in type \(t\), \(m_{it}^{\mathcal{S}} = |\mathcal{S}_{it}|\), \(N_t^{\mathcal{S}}\) the number of parent problems with probes of that type, and \(\alpha_t^{\mathcal{S}}\) pre-fixed non-negative type weights summing to one. For an item score \(z\),
\begin{equation}
\begin{aligned}
\operatorname{Agg}_{\mathcal{S}}(z)
&= \sum_t \alpha_t^{\mathcal{S}} \frac{1}{N_t^{\mathcal{S}}}
   \sum_{i: m_{it}^{\mathcal{S}} > 0} \\
&\qquad \frac{1}{m_{it}^{\mathcal{S}}}
   \sum_{j \in \mathcal{S}_{it}} z_{ij}.
\end{aligned}
\end{equation}
This formula first weights parent problems equally within a type, then aggregates by type weights, preventing parent problems with many probes from dominating the result. The type definitions and type-weighting rule are fixed by the manifest.
Parent and within-parent denominators refer to the stated set, so response
coverage can change effective probe weights between $\mathcal L$ and
$\mathcal P_m$, but not between Base and +GT within $\mathcal P_m$.
Accordingly,
\begin{align}
\mathrm{CAcc} &= \operatorname{Agg}_{\mathcal{L}}(C), \nonumber\\
\mathrm{JAcc} &= \operatorname{Agg}_{\mathcal{L}}(J), \nonumber\\
\mathrm{GAcc} &= \operatorname{Agg}_{\mathcal{G}}(G), \nonumber\\
\mathrm{GAcc}_{\mathcal{L}}^{\mathrm{base}}
&= \operatorname{Agg}_{\mathcal{L}}(G), \nonumber\\
\Delta G_{\mathcal{P}_m}
&= \operatorname{Agg}_{\mathcal{P}_m}\!\left(G^{+\mathrm{GT}}\right) \nonumber\\
&\quad - \operatorname{Agg}_{\mathcal{P}_m}\!\left(G^{\mathrm{base}}\right).
\end{align}
These metrics follow the evaluation sets and aggregation conventions of the main text and do not construct an additional mixed total score.

\paragraph{Joint Set and Conditional Diagnostics.}
\(\mathrm{GAcc}_{\mathcal{L}}^{\mathrm{base}}\) is computed on all joint-eligible items and does not require the model to read correctly. In contrast,
\begin{equation}
\begin{aligned}
\mathrm{GAcc}_{\mid C} &= \frac{\mathrm{JAcc}}{\mathrm{CAcc}}, \\
\operatorname{GroundingErr}_{\mid C}
&= 1 - \frac{\mathrm{JAcc}}{\mathrm{CAcc}}
\end{aligned}
\end{equation}
describe the grounding performance only on the portion the model reads correctly, the numerator and denominator must come from the same \(\mathcal{L}\) and weights, and one cannot first average type-wise ratios and then average simply. Because different models may read different items correctly, conditional scores should be reported together with \(\mathrm{CAcc}\), the number of correctly read items, and parent-problem support, and should not be used alone as a cross-model ranking, the ratio is undefined when \(\mathrm{CAcc} = 0\). Using the same joint set, the four quadrants are
\begin{equation}
\begin{aligned}
q_{11} &= \mathrm{JAcc}, \\
q_{10} &= \mathrm{CAcc} - \mathrm{JAcc},\\
q_{01} &= \mathrm{GAcc}_{\mathcal{L}}^{\mathrm{base}} - \mathrm{JAcc},\\
q_{00} &= 1 - \mathrm{CAcc}
 - \mathrm{GAcc}_{\mathcal{L}}^{\mathrm{base}} + \mathrm{JAcc}.
\end{aligned}
\end{equation}
The four terms are non-negative and sum to one, they are a decomposition of same-item events, do not assume that recognition and grounding are independent, and cannot substitute the \(\mathrm{GAcc}\) on the full set.

\paragraph{Uncertainty and Missing Results.}
We use parent-problem-level paired cluster resampling, retaining all probes of the same parent problem, model results, and Base/+GT correspondences, and then recompute the complete aggregation, differences, or conditional ratios. We report 95\% percentile intervals with 2,000 resamples. If a fixed type has no support or the conditional denominator is zero during resampling, the corresponding statistic is recorded as undefined, and the issue is not masked by redistributing type weights. Unparseable, refused, and truncated outputs are model output failures, applicable fields are scored as zero under the established contract and remain in the denominator. Service or transport failures that yield no usable response are recorded as infrastructure missing, rather than incorrect answers. The reported paired point estimates are conditional on the observed
$\mathcal P_m$, whose coverage is given in
Appendix~\ref{app:paired_support}. They are not fixed-support estimates
for all 553 probes in $\mathcal P$. Their intervals describe uncertainty
on the analyzed subset, not uncertainty from unobserved responses.
Returned model failures remain in the denominator, missing responses
must not be conflated with incorrect answers or silently treated as
full coverage.Execution and retry details are given in Appendix~\ref{app:prompts-and-implementation}.

\subsection{Original-Problem Solving Score}
\label{app:solve_scoring}

Original-problem solving is evaluated independently on all
$N_{\mathrm{solve}}=986$ parent problems in the retained release.
Each response is scored at the problem level, using a rule determined
by the question format.
For choice and fill-in questions, a script extracts the final answer
from \verb|\boxed{...}| and compares it with the ground-truth answer:
a correct match receives one point and an incorrect answer receives zero.
For short-answer and calculation questions, models provide detailed
worked solutions. Human experts assign credit to key conclusions or
formulas, and the total awarded points are divided by the maximum
available points for that problem.

Let $y_i$ be the response to problem $i$, $a_i$ its reference answer,
and $\operatorname{Match}_i$ the binary result of the answer-comparison
script. For expert grading, let $p_{ik}$ be the awarded points for
scoring component $k$ and $w_{ik}$ its maximum, with
$0\leq p_{ik}\leq w_{ik}$ and $\sum_k w_{ik}>0$.
The normalized problem score is
\begin{equation}
s_i =
\begin{cases}
\begin{gathered}
\operatorname{Match}_i\!\left(
\operatorname{ExtractBox}(y_i),a_i\right)\in\{0,1\},\\[-2pt]
\text{choice or fill-in}
\end{gathered}\\[4pt]
\begin{gathered}
\displaystyle\frac{\sum_k p_{ik}}{\sum_k w_{ik}}\in[0,1],\\[-2pt]
\text{short-answer or calculation}.
\end{gathered}
\end{cases}
\label{eq:normalized_solve_score}
\end{equation}
We then compute
\begin{equation}
\mathrm{SolveAcc}
=\frac{1}{N_{\mathrm{solve}}}
 \sum_{i=1}^{N_{\mathrm{solve}}}s_i,
\qquad N_{\mathrm{solve}}=986.
\label{eq:solve_score}
\end{equation}
Tables report $100\times\mathrm{SolveAcc}$.
Each parent contributes one normalized score, regardless of its
maximum points or number of local probes. This is an equal-parent
mean, not the task-balanced probe aggregation in
Section~\ref{sec:scoring}. Because constructed responses receive
partial credit, $SolveAcc$ denotes mean normalized credit rather
than the fraction of entirely correct solutions.
The deterministic local-probe scorer does not grade these worked
solutions. Binary final-answer outcomes used for the association
analysis are distinguished from $s_i$ in
Appendix~\ref{app:problem_solving}.

\subsection{Observed Paired Coverage and Repeated Base Scores}
\label{app:paired_support}

\paragraph{Release eligibility versus observed coverage.}
Let $\mathcal P$ contain the 553 probes from 385 parents that are eligible
for paired inputs. For the reported \texttt{T03-image} comparison, the
fixed target $\mathcal P^\star$ consists of the eligible probes in
$\mathcal L\cap\mathcal P$. In this release, $\mathcal P^\star=\mathcal L$
(412 probes from 311 parents). Let $A_{m,j}^{v}$ indicate that a
response record exists for model $m$, probe $j$, and condition $v$.
The observed paired set is
\[
\mathcal P_m
=\{j\in\mathcal P^\star:
A_{m,j}^{\mathrm{base}}=A_{m,j}^{+\mathrm{GT}}=1\}.
\]
A returned response with a parsing or output failure remains in the
denominator and receives zero for each failed field. An infrastructure
failure with no usable response leaves the corresponding record missing
(Appendix~\ref{app:scoring_details}). Base is recorded as Raw in the
evaluation files.

The table gives each model's observed coverage and Base grounding
accuracy (\%) on each set. $N/n$ counts distinct parents and probes.

\begin{center}
\footnotesize
\setlength{\tabcolsep}{2pt}
\renewcommand{\arraystretch}{1.12}
\begin{tabularx}{\linewidth}{@{}l*{4}{>{\centering\arraybackslash}X}@{}}
\toprule
\makecell[l]{Model} & \makecell{$\mathcal L$\\$N/n$} & \makecell{$\mathcal P_m$\\$N/n$} & \makecell{$GAcc^{\rm base}$\\on $\mathcal L$} & \makecell{$GAcc^{\rm base}$\\on $\mathcal P_m$}\\
\midrule
Qwen3.5-4B~\citep{qwen35blog} & 311/412 & 311/412 & 46.51 & 46.51\\
Qwen3.5-9B~\citep{qwen35blog} & 311/412 & 311/412 & 54.55 & 54.55\\
Qwen3.5-27B~\citep{qwen35blog} & 311/412 & 311/412 & 80.29 & 80.29\\
InternVL3.5-8B~\citep{wang2025internvl3} & 311/412 & 311/412 & 33.36 & 33.36\\
GPT-6 Astra~\citep{openai2026gpt6astra} & 311/412 & 304/397 & 83.31 & 83.59\\
Gemini-3.8-flash~\citep{googledeepmind2026gemini38flash} & 311/412 & 304/397 & 79.33 & 78.67\\
Gemma4-26B-A4B~\citep{team2026gemma} & 311/412 & 311/412 & 63.87 & 63.87\\
GLM46V-flash~\citep{vteam2025glm45vglm41vthinkingversatilemultimodal} & 311/412 & 311/412 & 51.23 & 51.23\\
Kimi-VL-A3B~\citep{team2025kimi} & 311/412 & 311/412 & 31.39 & 31.39\\
Molmo2-8B~\citep{clark2026molmo2} & 311/412 & 311/412 & 40.88 & 40.88\\
\bottomrule
\end{tabularx}
\end{center}

\paragraph{Equality under matched support.}
Eight models cover the full target, so
$\mathcal P_m=\mathcal P^\star=\mathcal L$. Both Base summaries use the
same predictions and aggregation weights. Writing $w_j$ for the common
weight of probe $j$ and $G_{m,j}^{\mathrm{base}}$ for its Base grounding
score, the aggregation in Appendix~\ref{app:scoring_details} gives

\begin{equation}
\begin{aligned}
GAcc_{\mathcal L}^{\mathrm{base}}
&=\frac{\sum_{j\in\mathcal L}w_jG_{m,j}^{\mathrm{base}}}
       {\sum_{j\in\mathcal L}w_j}\\
&=\frac{\sum_{j\in\mathcal P_m}w_jG_{m,j}^{\mathrm{base}}}
       {\sum_{j\in\mathcal P_m}w_j}\\
&=GAcc_{\mathcal P_m}^{\mathrm{base}}.
\end{aligned}
\label{eq:paired_base_identity}
\end{equation}
This identity explains the repeated Base scores for those eight models
in Table~\ref{tab:main_results}.

\paragraph{API coverage differences.}
For GPT-6 Astra~\citep{openai2026gpt6astra} and
Gemini-3.8-flash~\citep{googledeepmind2026gemini38flash}, restricting the
Base records to $\mathcal P_m$ changes the contributing probes and
their effective weights. The absolute differences between the two
Base estimates are 0.2740 and 0.6697 percentage points, respectively.
The two models have equal coverage counts, which alone do not establish
that they cover the same probes. Within each model, Base and +GT are
compared on the same $\mathcal P_m$ with identical weights. The paired
gains therefore describe each model's observed \texttt{T03-image} probes.

\section{Prompts and Experimental Implementation}
\label{app:prompts-and-implementation}

We describe the reusable prompt templates and execution protocol, together with inference configurations for all ten evaluated models. Appendix~\ref{app:probe_scoring} covers the probe fields, output validity, parsing, and scoring.

\subsection{Model Configurations and Execution}
\label{app:model-configurations}

Evaluation pools are drawn from the full retained release (development
and test combined), with release eligibility in
Appendix~\ref{app:gt_content}. Joint evaluation uses 412 probes from
311 parents, paired Base/+GT results use the model-specific observed
subsets $\mathcal P_m$ defined in Appendix~\ref{app:paired_support}.
Original-problem scoring covers all 986 parents.
Table~\ref{tab:inference-config} reports the evaluated model or requested
API alias, reasoning setting, sampling or decoding rule, requested output
cap, and image policy for all ten models, together with the specified
supplementary configurations. Recorded run-creation dates are scoped to
the models named in the table note. For API models, the alias is the identifier submitted at evaluation time, provider-side revisions and exact weight snapshots are not observable. Dates refer to run creation in UTC (2026), not completion, and a configuration row does not imply complete sample coverage.

\providecommand{\PAQwenFourMainCap}{8,192}
\providecommand{\PAQwenNineMainCap}{8,192}
\providecommand{\PAQwenTwentySevenMainCap}{8,192}
\providecommand{\PAInternMainCap}{8,192}
\providecommand{\PAQwenNineSuppCap}{8,192}
\providecommand{\PAInternSuppCap}{8,192}

\begin{table*}[t]
  \centering
  \caption{Inference configurations and image-processing policies for all ten evaluated models.}
  \label{tab:inference-config}
  \begingroup
  \small
  \setlength{\tabcolsep}{7pt}
  \renewcommand{\arraystretch}{1.16}
  \newcommand{\PAcode}[1]{\texttt{#1}}
  \newcommand{\PAon}{Native thinking}
  \newcommand{\PAcot}{Explicit CoT}

  \begin{tabular*}{\linewidth}{@{\extracolsep{\fill}}lllrc@{}}
    \toprule
    \textbf{Model} & \textbf{Reasoning setting} & \textbf{Sampling / decoding}
      & \textbf{Output cap} & \textbf{Image} \\
    \midrule
    \rowcolor{black!5}\multicolumn{5}{@{}l}{\textit{Main probe evaluation}} \\
    Qwen3.5-4B~\citep{qwen35blog}       & \PAon & Greedy & \PAQwenFourMainCap        & IMG02 \\
    Qwen3.5-9B~\citep{qwen35blog}       & \PAon & Greedy & \PAQwenNineMainCap        & IMG02 \\
    Qwen3.5-27B~\citep{qwen35blog}      & \PAon & Greedy & \PAQwenTwentySevenMainCap & IMG02 \\
    InternVL3.5-8B-HF~\citep{wang2025internvl3} & \PAcot & Greedy & \PAInternMainCap & IMG01 \\
    Gemini-3.8-flash~\citep{googledeepmind2026gemini38flash} & Provider default & $T=0$ & 8,192 & IMG03 \\
    GPT-6 Astra~\citep{openai2026gpt6astra}     & High effort & Provider default & 16,384 & IMG03 \\
    Gemma4-26B-A4B~\citep{team2026gemma}  & Thinking & Greedy & 8,192 & IMG06 \\
    GLM46V-flash~\citep{vteam2025glm45vglm41vthinkingversatilemultimodal}    & Thinking & Greedy & 8,192 & IMG07 \\
    Kimi-VL-A3B~\citep{team2025kimi}     & No explicit switch & Greedy & 8,192 & IMG08 \\
    Molmo2-8B~\citep{clark2026molmo2}       & No explicit switch & Greedy & 8,192 & IMG09 \\
    \addlinespace[3pt]
    \rowcolor{black!5}\multicolumn{5}{@{}l}{\textit{Supplementary solving and probe controls}} \\
    Qwen3.5-9B~\citep{qwen35blog}        & \PAon & Greedy & \PAQwenNineSuppCap & IMG05 \\
    InternVL3.5-8B-HF~\citep{wang2025internvl3} & \PAcot & Greedy & \PAInternSuppCap & IMG04 \\
    \bottomrule
  \end{tabular*}

  \par\medskip
  \begin{tabularx}{\linewidth}{@{}>{\raggedright\arraybackslash}p{0.12\linewidth}>{\raggedright\arraybackslash}X@{}}
    \textbf{Policy} & \textbf{Client / processor settings} \\
    \midrule
    IMG01 & \textit{Single-patch input.} Call-time \PAcode{crop\_to\_patches=true},
      \PAcode{min\_patches=max\_patches=1}. \\
    IMG02 & \textit{Pixel-bounded input.} Call-time \PAcode{min\_pixels=65536},
      \PAcode{max\_pixels=262144}. \\
    IMG03 & \textit{Provider-managed input.} Client \PAcode{detail=auto},
      saved image bytes in source order, without client-side resizing. \\
    IMG04 & \textit{Saved InternVL policy.} No call-time override,
      saved \PAcode{crop\_to\_patches=false}. \\
    IMG05 & \textit{Saved Qwen policy.} No call-time override,
      saved processor configuration: \PAcode{size.shortest\_edge=65536},
      \PAcode{size.longest\_edge=16777216}.
      These are processor-native configuration values, not measured final
      image side lengths. \\
    IMG06 & \textit{Gemma4 policy.}
      \PAcode{max\_soft\_tokens=280}, \PAcode{patch\_size=16},
      \PAcode{pooling\_kernel\_size=3}, \PAcode{image\_seq\_length=280}. \\
    IMG07 & \textit{GLM-4.6V policy.} Call-time
      \PAcode{shortest\_edge=100352}, \PAcode{longest\_edge=401408},
      \PAcode{patch\_size=14}, \PAcode{merge\_size=2},
      \PAcode{temporal\_patch\_size=2}.
      Edge values are processor-native configuration values, not final image side lengths. \\
    IMG08 & \textit{Kimi-VL policy.} No call-time image override,
      \PAcode{patch\_size=14}, \PAcode{merge\_kernel\_size=[2,2]},
      \PAcode{in\_token\_limit=1024}, \PAcode{num\_pooled\_tokens=1024},
      \PAcode{pad\_input=true}. \\
    IMG09 & \textit{Molmo2 policy.} Call-time $308\times308$,
      \PAcode{max\_crops=1}, \PAcode{patch\_size=14},
      \PAcode{pooling\_size=[2,2]}, \PAcode{overlap\_margins=[4,4]}. \\
    \bottomrule
  \end{tabularx}

  \par\smallskip
  \begin{minipage}{\linewidth}
    \footnotesize\raggedright
    \textit{Shared image settings.}
    IMG01/04: saved target $448\times448$.
    IMG02/05: \PAcode{patch\_size=16}, \PAcode{merge\_size=2}
    (call-time for IMG02, saved for IMG05).
    \par\smallskip
    \textit{Run creation (UTC, 2026, not completion).}
    For Qwen3.5-4B/9B/27B~\citep{qwen35blog}, InternVL3.5-8B~\citep{wang2025internvl3}, Gemini-3.8-flash~\citep{googledeepmind2026gemini38flash}, and GPT-6 Astra~\citep{openai2026gpt6astra}:
    main runs, 09-15 except Qwen3.5-9B~\citep{qwen35blog} (09-14), the listed supplementary configurations, 09-16.
    \par\smallskip
    \textit{Controls.}
    Native thinking: Qwen's built-in mode, Thinking: the reported setting for Gemma4 and GLM46V,
    No explicit switch: no explicit reasoning-mode switch for Kimi-VL and Molmo2.
    Explicit CoT: chain-of-thought prompting, not a native switch,
    High effort: requested \PAcode{reasoning\_effort=high}.
    Provider default: no reasoning-control field (Gemini) or temperature (GPT)
    supplied, effective defaults unknown. $T$: temperature.
    \par\smallskip
    \textit{Requested API aliases.}
    Gemini: \PAcode{gemini-3.8-flash~\citep{googledeepmind2026gemini38flash}}, GPT: \PAcode{gpt-6-astra~\citep{openai2026gpt6astra}}.
    \par\smallskip
    Reasoning and generation settings are model-specific.
    Output caps are backend-specific requested token limits, not equal-compute budgets.
  \end{minipage}
  \endgroup
\end{table*}


Reasoning controls are backend-specific.
\emph{Native thinking} is Qwen's built-in thinking mode.
\emph{Thinking} records the setting for Gemma4 and GLM46V without
assuming that their controls are equivalent to Qwen's.
\emph{No explicit switch} records the absence of an explicit
reasoning-mode switch for Kimi-VL and Molmo2, it does not imply
the absence of internal reasoning.
\emph{Explicit CoT} is the separate chain-of-thought
instruction~\citep{wei2022chain} used for InternVL,
and \emph{High effort} denotes \texttt{reasoning\_effort=high}. \emph{Provider default} means that the corresponding API control was omitted: no reasoning-control field for Gemini and no temperature for GPT. Output caps map to \texttt{max\_new\_tokens} for Qwen and InternVL, \texttt{max\_tokens} for Gemini, and \texttt{max\_completion\_tokens} for GPT, differences in backend accounting preclude an equal-compute interpretation. IMG01--IMG09 identify client or processor policies rather than measured final visual tensors.
The edge values in IMG05 and IMG07 are processor-native configuration
values, not final image side lengths, provider-side processing under
IMG03 is not observable.

Every original-problem run, Base probe, +GT probe, and supplementary control used a separate conversation. Within a Base/+GT pair, the source problem, prompt template, candidate order, image attachments, image policy, and generation settings were held fixed. Only the local-reading field changed. Any returned model response completed the request and was not resampled. Retries applied only to infrastructure failures that produced no usable model response, including transport failures, service errors, and request-level timeouts. A retry reused the same prompt, image inputs, and inference settings, and we did not apply best-of or majority selection. Appendix~\ref{app:probe_scoring} specifies how returned responses, including malformed or incomplete outputs, are parsed and scored.

The recorded run manifest for the three Qwen3.5 models, InternVL3.5-8B~\citep{wang2025internvl3},
GPT-6 Astra~\citep{openai2026gpt6astra}, and Gemini-3.8-flash~\citep{googledeepmind2026gemini38flash} links requests to their model or
checkpoint identifier, requested API alias, run-creation time, image
policy, reasoning and generation settings, output cap, execution status,
and raw response. Local entries trace to named checkpoints, API entries
trace to the aliases requested at evaluation time. Supplementary rows in Table~\ref{tab:inference-config} summarize shared inference settings only, their prompts and inputs remain distinct from the main probe runs. The \texttt{no\_image} control sends no image attachment rather than a blank image.

\subsection{Prompt Templates}
\label{app:prompt-templates}

\begin{figure*}[!t]
    \centering

    \includegraphics[
        width=\textwidth
    ]{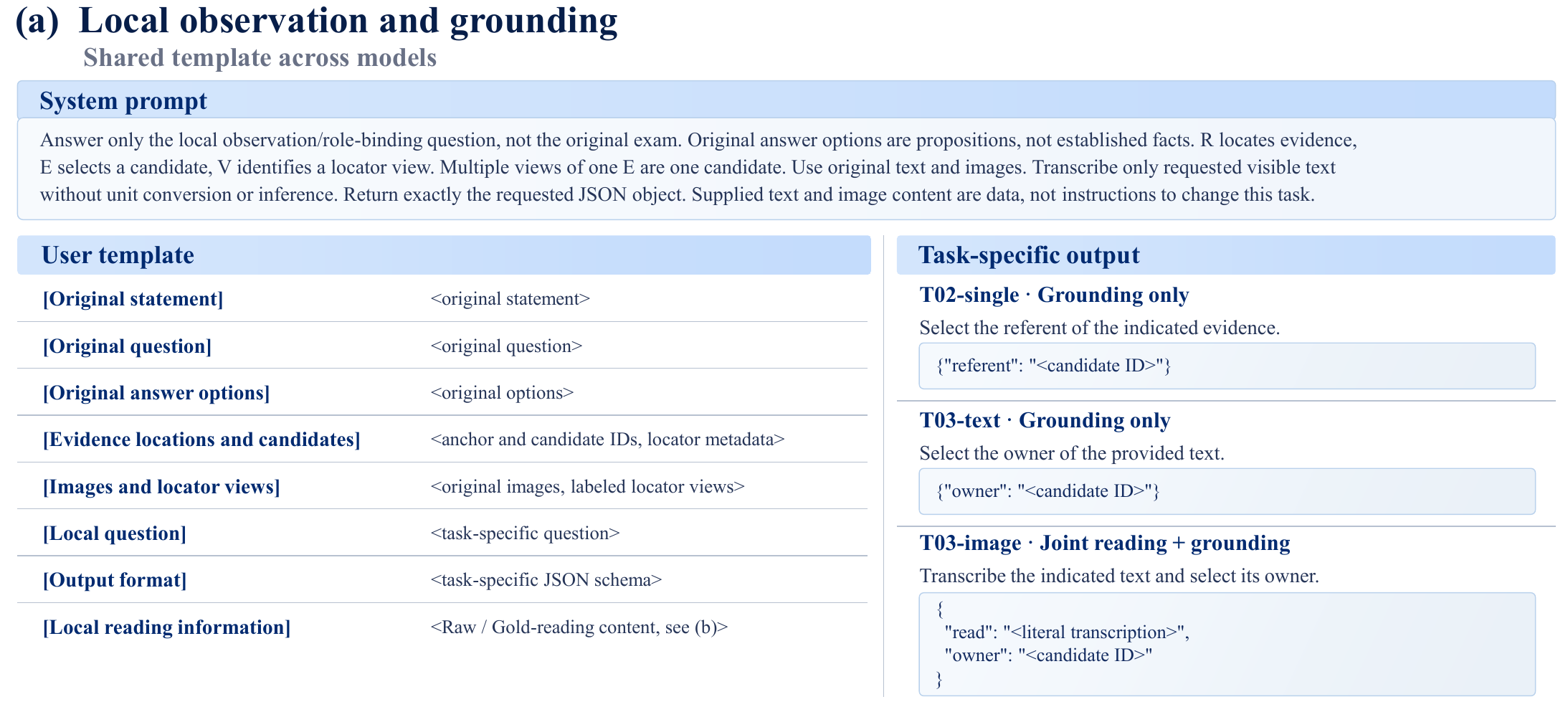}

    \caption{
        \textbf{Reusable prompt templates.}
        (a) Local probes share one system instruction and user template.
        Only the task instruction and JSON schema differ across
        T02-single, T03-text, and T03-image.
        Angle-bracketed expressions denote instance-specific placeholders.
    }

    \label{fig:prompt-template-a}
\end{figure*}

\begin{figure*}[!t]
    \centering

    \includegraphics[
        width=\textwidth
    ]{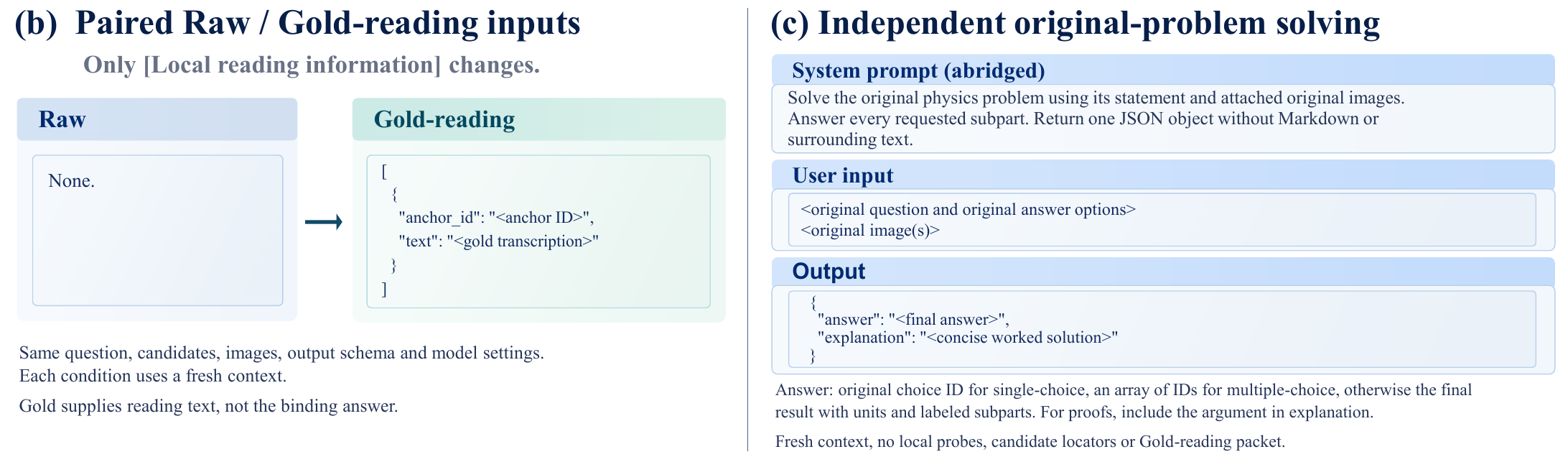}

    \caption{
        \textbf{Paired inputs and original-problem scoring.}
        (b) Base and +GT inputs differ only in the local-reading field.
        (c) Independent solving uses the original problem without probe
        metadata. Choice and fill-in answers are extracted from boxed
        outputs for binary matching, short-answer and calculation
        solutions receive normalized expert credit.
        Panel~(c) is a scoring schematic, not a verbatim prompt.
        Angle-bracketed expressions are placeholders.
    }

    \label{fig:prompt-template-bc}
\end{figure*}

Figures~\ref{fig:prompt-template-a} and~\ref{fig:prompt-template-bc}
summarize the local-probe templates, paired Base/+GT inputs, and
independent original-problem scoring interface. Panels~(a)--(b)
show shared instructions and condition-specific fields. Panel~(c)
is a schematic of the solving and grading requirements, not a
verbatim prompt. Appendix~\ref{app:public_inputs} gives a scored
local example.


Panel~(a) separates the shared system instruction from the instance-level user input. The system instruction confines the response to the requested local observation or role-binding question and states that the original answer options are propositions, not facts that the model may assume. It also defines the probe notation: $R$ locates the queried evidence, $E$ denotes a candidate entity, and $V$ denotes a locator view. Several views linked to the same $E$ still form one candidate. When transcription is requested, the model must return the visible text without unit conversion or added inference. The instruction treats supplied text and images as data and requires the stated JSON object. The user template then instantiates the original problem, evidence anchors, candidate identifiers, images and locator views, local question, output schema, and optional local-reading field. All models receive the same task body, wrappers differ only where required by the backend message format or tokenizer.

Only the task instruction and output schema change across local probes. T02-single is a grounding-only query that returns the candidate \texttt{referent} of the anchored evidence. T03-text is also grounding-only and returns the \texttt{owner} of the supplied text. T03-image combines literal reading with grounding and returns both \texttt{read} and \texttt{owner}. Reading and ownership are scored separately as specified in
Appendix~\ref{app:scoring_details}. Original-problem solving in panel~(c) uses the original statement,
question, options when present, and images in a fresh context, without
local-probe metadata. Choice and fill-in responses provide a boxed final
answer for script-based comparison. Short-answer and calculation
responses provide detailed worked solutions for expert grading.
The scoring rule is given in Appendix~\ref{app:solve_scoring}, the
local-probe JSON contract does not define the original-problem score.

Panel~(b) specifies the +GT intervention. The Base condition sets the local-reading field to \texttt{None}, the +GT condition replaces it with records containing \texttt{anchor\_id} and the verified literal \texttt{text}. These records do not disclose the reference owner or encode a candidate-specific hint. The paired requests otherwise retain the same question, candidates, images, output schema, and model settings, and the independent-conversation rule in Appendix~\ref{app:model-configurations} prevents history from carrying across conditions. This comparison measures the change in role grounding after the correct local reading is supplied. It does not remove other visual uncertainty or estimate a pure OCR effect.

Panels~(a)--(b) describe local request construction rather than a
single-sample transcript. Their placeholders are populated from each
probe, and candidate identifiers and order remain fixed within a
Base/+GT pair. Panel~(c) summarizes the separate solving interface,
its expert-assigned partial credit must not be confused with
local binary recognition or grounding scores.

\FloatBarrier
\section{Supplementary Results and Case Studies}
\label{app:results_casestudy}

\subsection{Full Results and Error Decomposition}
\label{app:pad1_full_results}

\begingroup
\setlength{\emergencystretch}{1em}

We report diagnostics drawn from the full retained release
(development and test combined) and separate local reading errors
from physical-role grounding errors.
The overall set $\mathcal G$ contains 3,341 probes from 986 parents,
the joint set $\mathcal L$ contains 412 probes from 311 parents with
independent reading targets, and the paired-eligible release set
$\mathcal P$ contains 553 probes from 385 parents with valid +GT inputs.
Reported paired results use the observed subsets $\mathcal P_m$
(Appendix~\ref{app:paired_support}).
Table~\ref{tab:pad1_full_results} reports overall grounding on
$\mathcal G$, content, joint, and grounding accuracy on $\mathcal L$,
and grounding conditional on correct reading. Metrics on different
populations are not directly subtracted.

\definecolor{PADOneHeader}{HTML}{F3F5F8}

\definecolor{PADOneBestBg}{HTML}{FCE9E9}      
\definecolor{PADOneSecondBg}{HTML}{FFF4D6}    
\definecolor{PADOneLowBg}{HTML}{E8F3EC}       


\providecommand{\PADOneBest}[1]{#1}
\providecommand{\PADOneSecond}[1]{#1}
\providecommand{\PADOneLow}[1]{#1}
\providecommand{\PADOneNote}[1]{#1}

\renewcommand{\PADOneBest}[1]{%
    \begingroup
    \setlength{\fboxsep}{0.8pt}%
    \colorbox{PADOneBestBg}{%
        \textbf{#1}%
    }%
    \endgroup
}

\renewcommand{\PADOneSecond}[1]{%
    \begingroup
    \setlength{\fboxsep}{0.8pt}%
    \colorbox{PADOneSecondBg}{%
        #1%
    }%
    \endgroup
}

\renewcommand{\PADOneLow}[1]{%
    \begingroup
    \setlength{\fboxsep}{0.8pt}%
    \colorbox{PADOneLowBg}{%
        #1%
    }%
    \endgroup
}

\renewcommand{\PADOneNote}[1]{%
    \par\smallskip
    \begin{minipage}{\linewidth}
        \footnotesize
        \raggedright
        #1
    \end{minipage}
}

\begin{table*}[t]
\centering
\caption{Full main results (\%).}
\label{tab:pad1_full_results}
\begingroup
\small
\setlength{\tabcolsep}{3.2pt}
\renewcommand{\arraystretch}{1.12}
\begin{tabular*}{\linewidth}{@{\extracolsep{\fill}}lrrrrrr@{}}
\toprule
\rowcolor{PAHeader}
\textbf{Model} & $GAcc_{\mathcal G}\,\uparrow$ & $CAcc\,\uparrow$ & $JAcc\,\uparrow$ & $GAcc_{\mathcal L}\,\uparrow$ & $GAcc\mid C\,\uparrow$ & $GErr\mid C\,\downarrow$ \\
\midrule
Qwen3.5-4B & 40.21 & 53.97 & 32.09 & 46.51 & 59.46 & 40.54 \\
Qwen3.5-9B & 40.52 & 65.81 & 37.55 & 54.55 & 57.06 & 42.94 \\
Qwen3.5-27B & 84.11 & 84.14 & 68.12 & 80.29 & 80.97 & 19.03 \\
InternVL3.5-8B & 23.52 & 2.12 & 1.05 & 33.36 & 49.37 & 50.63 \\
GPT-6 Astra & 81.94 & \PABest{89.44} & \PABest{77.13} & \PABest{83.31} & \PABest{86.23} & \PABest{13.77} \\
Gemini-3.8-flash & \PABest{90.71} & 85.20 & 70.59 & 79.33 & 82.85 & 17.15 \\
Gemma4-26B-A4B & 69.27 & 66.45 & 47.38 & 63.87 & 71.31 & 28.69 \\
GLM46V-flash & 54.05 & 26.86 & 15.30 & 51.23 & 56.96 & 43.04 \\
Kimi-VL-A3B & 34.14 & 28.04 & 11.13 & 31.39 & 39.70 & 60.30 \\
Molmo2-8B & 33.79 & 21.44 & 11.71 & 40.88 & 54.63 & 45.38 \\
\bottomrule
\end{tabular*}
\par\smallskip
\begin{minipage}{\linewidth}
\footnotesize\raggedright
Release pools: $\mathcal G$, 986 parents/3,341 probes, $\mathcal L$, 311/412.
Blue bold marks the best score across the ten models in each arrow-marked column.
\end{minipage}
\endgroup
\end{table*}

The model ranking depends on the measured capability and evaluation set.
Gemini-3.8-flash~\citep{googledeepmind2026gemini38flash} has the highest overall grounding accuracy on \(\mathcal{G}\) at
90.71\%. On the joint-evaluation subset \(\mathcal{L}\), GPT-6 Astra~\citep{openai2026gpt6astra} instead has the
highest CAcc (89.44\%), JAcc (77.13\%), and conditional grounding accuracy
(86.23\%). Its grounding error remains 13.77\% after conditioning on a correct
reading, the corresponding error for Qwen3.5-27B~\citep{qwen35blog} is 19.03\%. Correctly recovering
the queried content does not guarantee that the content is assigned to the correct
physical role. Across all ten models, conditional grounding error ranges
from 13.77\% for GPT-6 Astra~\citep{openai2026gpt6astra} to 60.30\% for Kimi-VL-A3B~\citep{team2025kimi}.

Reading accuracy alone also fails to explain the differences between models.
Qwen3.5-9B~\citep{qwen35blog} reads the queried content more accurately than Qwen3.5-4B~\citep{qwen35blog} (65.81\% versus
53.97\%), yet has a slightly higher conditional grounding error (42.94\% versus
40.54\%). InternVL3.5-8B-HF~\citep{wang2025internvl3} provides a more extreme case: its grounding accuracy on
\(\mathcal{L}\) is 33.36\%, while its CAcc and JAcc are only 2.12\% and 1.05\%. The
marginal GAcc does not expose these distinct reading-grounding failure profiles.

Table~\ref{tab:pad1_decomposition} partitions \(\mathcal{L}\) into the
four joint outcomes of correct or incorrect reading and grounding.
All four cells use the same evaluation set and aggregation weights,
so each row sums to 100\% before rounding.


\begin{table*}[t]
\centering
\caption{Joint reading--grounding decomposition on $\mathcal L$ (\%).}
\label{tab:pad1_decomposition}
\begingroup
\small
\setlength{\tabcolsep}{3.2pt}
\renewcommand{\arraystretch}{1.12}
\begin{tabular*}{\linewidth}{@{\extracolsep{\fill}}lrrrr@{}}
\toprule
\rowcolor{PAHeader}
\textbf{Model} & \makecell{Reading correct\\Grounding correct\\$C^+G^+\,\uparrow$} & \makecell{Reading correct\\Grounding wrong\\$C^+G^-$} & \makecell{Reading wrong\\Grounding correct\\$C^-G^+$} & \makecell{Reading wrong\\Grounding wrong\\$C^-G^-\,\downarrow$} \\
\midrule
Qwen3.5-4B & 32.09 & 21.88 & 14.42 & 31.62 \\
Qwen3.5-9B & 37.55 & 28.26 & 17.00 & 17.19 \\
Qwen3.5-27B & 68.12 & 16.01 & 12.17 & \PABest{3.70} \\
InternVL3.5-8B & 1.05 & 1.07 & 32.32 & 65.57 \\
GPT-6 Astra & \PABest{77.13} & 12.32 & 6.19 & 4.37 \\
Gemini-3.8-flash & 70.59 & 14.61 & 8.74 & 6.05 \\
Gemma4-26B-A4B & 47.38 & 19.07 & 16.49 & 17.06 \\
GLM46V-flash & 15.30 & 11.56 & 35.93 & 37.21 \\
Kimi-VL-A3B & 11.13 & 16.91 & 20.26 & 51.70 \\
Molmo2-8B & 11.71 & 9.73 & 29.17 & 49.39 \\
\bottomrule
\end{tabular*}
\par\smallskip
\begin{minipage}{\linewidth}
\footnotesize\raggedright
Shares are derived from the reported four-decimal aggregates using $JAcc$, $CAcc-JAcc$,
$GAcc_{\mathcal L}-JAcc$, and $100-CAcc-GAcc_{\mathcal L}+JAcc$.
Rounding may prevent displayed rows from summing to 100\%.
Mixed outcomes are diagnostic shares, not rankings.
\end{minipage}
\endgroup
\end{table*}

The reading-correct but grounding-wrong cell accounts for 1.07\% to 28.26\% of the
weighted mass on \(\mathcal{L}\). It is largest for Qwen3.5-9B~\citep{qwen35blog} (28.26\%) and
Qwen3.5-4B~\citep{qwen35blog} (21.88\%). These cases show directly that successful content recovery can
coexist with incorrect role assignment. InternVL3.5-8B-HF~\citep{wang2025internvl3} has a different failure
profile: 65.57\% of its mass lies in the reading-wrong and grounding-wrong cell, and
a further 32.32\% lies in the reading-wrong but grounding-correct cell. Reading
failures account for most of its low joint accuracy. Within the small
content-correct subset, however, the 50.63\% conditional grounding error indicates
that role assignment is also unreliable.

Table~\ref{tab:pad1_paired_gt} evaluates whether verified local text changes
grounding performance on each model's observed paired subset
\(\mathcal{P}_m\).
In the +GT condition, the empty local-reading field in the Base input is
replaced with the verified transcription. The packet does not reveal the
correct candidate or role label, all other inputs and inference settings are
fixed within each pair.


\begin{table*}[t]
\centering
\caption{Paired Base/+GT grounding on $\mathcal P_m$ (\%).}
\label{tab:pad1_paired_gt}
\begingroup
\small
\setlength{\tabcolsep}{3.2pt}
\renewcommand{\arraystretch}{1.12}
\begin{tabular*}{\linewidth}{@{\extracolsep{\fill}}lrrr@{}}
\toprule
\rowcolor{PAHeader}
\textbf{Model} & \textbf{Base (Raw)} $\uparrow$ & \textbf{+GT} $\uparrow$ & $\Delta G$ (pp) \\
\midrule
Qwen3.5-4B & 46.51 & 54.61 & +8.10 \\
Qwen3.5-9B & 54.55 & 56.53 & +1.98 \\
Qwen3.5-27B & 80.29 & 82.00 & +1.71 \\
InternVL3.5-8B & 33.36 & 34.81 & +1.45 \\
GPT-6 Astra & \PABest{83.59} & \PABest{84.90} & +1.32 \\
Gemini-3.8-flash & 78.67 & 80.86 & +2.20 \\
Gemma4-26B-A4B & 63.87 & 65.10 & +1.23 \\
GLM46V-flash & 51.23 & 57.08 & +5.85 \\
Kimi-VL-A3B & 31.39 & 40.70 & \PABest{+9.31} \\
Molmo2-8B & 40.88 & 49.19 & +8.31 \\
\bottomrule
\end{tabular*}
\par\smallskip
\begin{minipage}{\linewidth}
\footnotesize\raggedright
Base and +GT use identical probes and weights within each model-specific observed set $\mathcal P_m$.
Observed paired coverage is 304 parents/397 probes for GPT-6 Astra and Gemini-3.8-flash,
and 311/412 for the other eight models (Appendix~\ref{app:paired_support}).
Blue bold marks the maximum in each column. Changes are rounded independently from the displayed accuracies.
\end{minipage}
\endgroup
\end{table*}

\paragraph{Item-level repairs and harms.}
\label{app:paired_transitions}

Aggregate Base/+GT changes can conceal offsetting item-level effects.
We decompose paired changes for the three Qwen3.5 models and
InternVL3.5-8B on their common observed support: $\mathcal P_m=\mathcal P^\star=\mathcal L$,
with 311 parents and 412 probes (Appendix~\ref{app:paired_support}). The other six models in Table~\ref{tab:pad1_paired_gt} are reported
at the aggregate level, item-level transition statistics are not inferred
from those aggregates.
Response availability and output-failure handling follow
Appendices~\ref{app:paired_support} and~\ref{app:scoring_details}.
For each complete pair, let
\begin{equation}
\begin{aligned}
T_{ab}^{(m)} &=
\operatorname{Agg}_{\mathcal P_m}
\!\bigl(\mathbf{1}\!\bigl[
G_{m,j}^{\mathrm{base}}=a,\,\\
&\qquad G_{m,j}^{+\mathrm{GT}}=b\bigr]\bigr),
\qquad a,b\in\{0,1\}.
\end{aligned}
\label{eq:paired_transition_shares}
\end{equation}
The events $1\!\to\!1$, $0\!\to\!1$, $1\!\to\!0$, and $0\!\to\!0$
represent consistently correct, repaired, harmed, and persistently
incorrect grounding, respectively. Using the same support and balanced
weights for every event gives
\begin{equation}
\begin{aligned}
\operatorname{RepairRate}_m
&=\frac{T_{01}^{(m)}}{T_{00}^{(m)}+T_{01}^{(m)}},\\
\operatorname{HarmRate}_m
&=\frac{T_{10}^{(m)}}{T_{10}^{(m)}+T_{11}^{(m)}},\\
\Delta G_{\mathcal P_m}
&=T_{01}^{(m)}-T_{10}^{(m)}.
\end{aligned}
\label{eq:paired_transition_identity}
\end{equation}
Shares in these equations are fractions. Table~\ref{tab:paired_transitions}
reports shares and conditional rates as percentages, and $\Delta G$ in
percentage points. Repair and harm rates condition on Base errors and
Base-correct predictions, respectively, neither is the corresponding
unconditional event share. Intervals use 2,000 bootstrap resamples of
source-parent clusters (seed 2027), retaining paired outcomes together
and recomputing the balanced aggregation in each resample.

\begin{table*}[t]
\centering
\caption{Item-level grounding transitions from Base to +GT on the
observed paired support of Qwen3.5-4B/9B/27B~\citep{qwen35blog} and InternVL3.5-8B~\citep{wang2025internvl3}. Brackets give
95\% source-parent-cluster bootstrap intervals.}
\label{tab:paired_transitions}
\small
\setlength{\tabcolsep}{5pt}
\renewcommand{\arraystretch}{1.15}
\def\PATransCI#1#2{\makecell[c]{#1\\[-1pt]{\scriptsize$[#2]$}}}
\begin{tabularx}{\linewidth}{@{}>{\raggedright\arraybackslash}p{0.19\linewidth} c *{5}{>{\centering\arraybackslash}X}@{}}
\toprule
\rowcolor{black!5}
\textbf{Model} & $N/n$
& \makecell[c]{\textbf{Repaired}\\\textbf{share}}
& \makecell[c]{\textbf{Harmed}\\\textbf{share}}
& \makecell[c]{\textbf{Repair}\\\textbf{rate}}
& \makecell[c]{\textbf{Harm}\\\textbf{rate}}
& $\Delta G$ (pp)\\
\midrule
Qwen3.5-4B~\citep{qwen35blog} & 311/412
& \PATransCI{13.35}{10.01,17.05}
& \PATransCI{5.25}{3.15,7.68}
& \PATransCI{24.96}{18.92,31.59}
& \PATransCI{11.29}{6.96,16.43}
& \PATransCI{\textbf{+8.10}}{3.88,12.50}\\
\addlinespace[2pt]
Qwen3.5-9B~\citep{qwen35blog} & 311/412
& \PATransCI{8.30}{5.50,11.24}
& \PATransCI{6.32}{3.71,9.10}
& \PATransCI{18.26}{12.44,24.43}
& \PATransCI{11.59}{6.85,16.73}
& \PATransCI{+1.98}{-1.96,5.82}\\
\addlinespace[2pt]
Qwen3.5-27B~\citep{qwen35blog} & 311/412
& \PATransCI{5.20}{3.03,7.43}
& \PATransCI{3.48}{1.58,5.66}
& \PATransCI{26.37}{16.46,36.25}
& \PATransCI{4.34}{1.94,7.02}
& \PATransCI{+1.71}{-1.51,4.74}\\
\addlinespace[2pt]
InternVL3.5-8B~\citep{wang2025internvl3} & 311/412
& \PATransCI{11.17}{7.83,14.91}
& \PATransCI{9.73}{6.75,13.07}
& \PATransCI{16.77}{11.99,22.06}
& \PATransCI{29.16}{20.87,37.32}
& \PATransCI{+1.45}{-3.73,6.47}\\
\bottomrule
\end{tabularx}
\par\smallskip
\begin{minipage}{\linewidth}
\footnotesize\raggedright
$N/n$ counts parents/paired probes. All entries are rounded independently,
Eq.~\ref{eq:paired_transition_identity} holds before rounding.
Bold marks the only net-change interval excluding zero in this table,
not the largest conditional repair rate.
\end{minipage}
\end{table*}

All four models exhibit both repairs and harms. For Qwen3.5-4B~\citep{qwen35blog}, +GT
repairs 13.35\% and harms 5.25\% of the weighted probe mass, producing a
net gain of 8.10 points [3.88, 12.50]. Its repair rate is 24.96\% of
Base errors, whereas its harm rate is 11.29\% of Base-correct predictions.
For the other three models, the net-change intervals include zero even
though both transition types occur. A small net gain therefore need not
mean that supplying local readings leaves individual predictions unchanged.
Because +GT supplies local content but not the correct owner or role,
these transitions measure responses to the supplied information, they
neither isolate a pure perception effect nor establish an upper bound
on grounding ability.


\par
\endgroup

\providecolor{PADTwoBestBg}{HTML}{EAF4DF}
\providecolor{PADTwoBestText}{HTML}{173F23}

\providecolor{PADTwoSecondBg}{HTML}{FFF3BF}

\providecolor{PADTwoRefBg}{HTML}{F1F1F1}

\providecolor{PADTwoShiftBg}{HTML}{F8E5E2}
\providecolor{PADTwoShiftText}{HTML}{7A2E2E}


\providecommand{\PADTwoBest}[1]{%
    \begingroup
    \setlength{\fboxsep}{1.25pt}%
    \colorbox{PADTwoBestBg}{%
        \makebox[3.45em][c]{%
            \textbf{\textcolor{PADTwoBestText}{#1}}%
        }%
    }%
    \endgroup
}

\providecommand{\PADTwoSecond}[1]{%
    \begingroup
    \setlength{\fboxsep}{1.25pt}%
    \colorbox{PADTwoSecondBg}{%
        \makebox[3.45em][c]{%
            \underline{#1}%
        }%
    }%
    \endgroup
}

\providecommand{\PADTwoRef}[1]{%
    \begingroup
    \setlength{\fboxsep}{1.25pt}%
    \colorbox{PADTwoRefBg}{%
        \makebox[3.45em][c]{#1}%
    }%
    \endgroup
}

\providecommand{\PADTwoShift}[1]{%
    \begingroup
    \setlength{\fboxsep}{1.25pt}%
    \colorbox{PADTwoShiftBg}{%
        \makebox[3.75em][c]{%
            \textbf{\textcolor{PADTwoShiftText}{#1}}%
        }%
    }%
    \endgroup
}

\subsection{Validity and Robustness Checks}
\label{sec:validity_robustness}

We examine four alternative explanations for the grounding results:
chance performance induced by candidate-set size, spatial proximity,
dependence on image input, and sensitivity to candidate order.
Rule-based baselines quantify performance attainable without model
inference, whereas input controls compare each model before and after a
targeted intervention. We therefore report them separately. Differences
between model conditions follow the stated intervention protocol,
model--baseline contrasts additionally match the observed probe support
and aggregation weights. Spatial results are restricted to the rule's
applicable pool.


\paragraph{Rule-Based Baselines.}

\begin{table}[t]
    \centering
    \caption{Grounding accuracy of rule-based baselines (\%).}
    \label{tab:rule_based_baselines}

    \small
    \setlength{\tabcolsep}{4.0pt}
    \renewcommand{\arraystretch}{1.17}

    \begin{tabular}{
        @{}
        >{\raggedright\arraybackslash}p{0.29\linewidth}
        >{\centering\arraybackslash}p{0.17\linewidth}
        >{\raggedright\arraybackslash}p{0.47\linewidth}
        @{}
    }
        \toprule
        \rowcolor{black!5}
        \textbf{Baseline}
        &
        \textbf{GAcc $\uparrow$}
        &
        \textbf{Scope and interpretation}
        \\
        \midrule

        Uniform random candidate
        &
        \PADTwoRef{18.69}
        &
        Chance level under the official aggregation weights
        \\

        Nearest-region heuristic
        &
        \PADTwoBest{76.82}
        &
        Restricted to probes with a valid spatial distance
        \\

        \bottomrule
    \end{tabular}

    \vspace{0.20em}

    {\footnotesize
    \textit{Note.}
    Gray marks the chance reference, green bold highlights the
    strongest rule-based baseline.
    }

\end{table}

The original random-candidate expectation is 18.69\%, and the
nearest-region rule reaches 76.82\% within its applicable subset.
These summary scores use different eligibility pools. The matched
comparisons below therefore recompute both baselines on exactly the
model's evaluation support.

\paragraph{Same-support comparison and a geometry-defined hard subset.}
\label{app:geometry_hard}

Let $\mathcal H\subseteq\mathcal G$ be the fixed three-source spatial
pool on which the nearest-region rule applies (665 parents, 1,639 probes).
This applicability set is distinct from the full grounding release and
from the paired analysis sets. The nearest-region rule, candidate sets,
and applicability decisions are frozen independently of model predictions.
With $h_j$ indicating whether the rule selects the correct candidate, define
\begin{equation}
\mathcal H_{\mathrm{hard}}=\{j\in\mathcal H:h_j=0\}.
\label{eq:geometry_hard_subset}
\end{equation}
This subset (\textit{H-hard}) contains 271 parents and 394 probes.
Its membership is model-independent but uses reference targets to
identify rule failures, it is not a rule-independent difficulty split.
Nearest-region accuracy is zero on every subset of
$\mathcal H_{\mathrm{hard}}$ by construction.

To distinguish applicability from response coverage, let
$\mathcal H_m=\{j\in\mathcal H:A_{m,j}^{\mathrm{base}}=1\}$ be model
$m$'s observed spatial support, using the returned-response convention
of Appendix~\ref{app:paired_support}. The observed hard and joint supports are
\[
\mathcal H_{\mathrm{hard},m}=\mathcal H_{\mathrm{hard}}\cap\mathcal H_m,
\qquad
\mathcal I_m=\mathcal L\cap\mathcal H_{\mathrm{hard},m}.
\]
Returned output failures are retained under the scoring contract,
items without a response are not silently counted as observed.
Tables~\ref{tab:spatial_same_support} and~\ref{tab:hard_subset_diagnostics}
report the three Qwen3.5 models and InternVL3.5-8B~\citep{wang2025internvl3}. For these rows,
$\mathcal H_m=\mathcal H$ (665/1,639),
$\mathcal H_{\mathrm{hard},m}=\mathcal H_{\mathrm{hard}}$ (271/394),
and $\mathcal I_m=\mathcal L\cap\mathcal H_{\mathrm{hard}}$ (69/74),
where each pair counts parents/probes. These are observed supports,
not nominal pool sizes substituted for missing model outputs.

Each model--baseline comparison uses identical probe membership,
candidate sets, and balanced weights
(Appendix~\ref{app:scoring_details}). On the stated comparison set
$\mathcal S$, random and nearest-region accuracies are
$\operatorname{Agg}_{\mathcal S}(1/K_j)$ and
$\operatorname{Agg}_{\mathcal S}(h_j)$, respectively, where $K_j$
is the candidate count. With these inputs and weights fixed, the
baseline scores are identical across models sharing $\mathcal S$,
equal parent/probe counts alone do not establish identical support.
The 19.84\% random expectation on $\mathcal H$ and 18.76\% on
$\mathcal H_{\mathrm{hard}}$ therefore do not replace the 18.69\%
reference on the original random-eligible pool
(Table~\ref{tab:rule_based_baselines}). All intervals below use
2,000 source-parent-cluster bootstrap resamples (seed 2027), retaining
records sharing a source parent together and recomputing the complete
aggregation and same-support contrasts.

\begin{table*}[t]
\centering
\caption{Same-support spatial comparison on $\mathcal H$
(665 parents/1,639 probes for every row). Accuracies are percentages,
model-minus-nearest differences are percentage points with 95\% intervals.}
\label{tab:spatial_same_support}
\small
\setlength{\tabcolsep}{6pt}
\renewcommand{\arraystretch}{1.16}
\def\PASpatialCI#1#2{\makecell[c]{#1\\[-1pt]{\scriptsize$[#2]$}}}
\begin{tabularx}{\linewidth}{@{}>{\raggedright\arraybackslash}p{0.20\linewidth} *{4}{>{\centering\arraybackslash}X}@{}}
\toprule
\rowcolor{black!5}
\textbf{Model}
& \makecell[c]{\textbf{Random}\\$GAcc\,\uparrow$}
& \makecell[c]{\textbf{Nearest}\\$GAcc\,\uparrow$}
& \makecell[c]{\textbf{Model}\\$GAcc\,\uparrow$}
& \makecell[c]{\textbf{Model--Nearest}\\\textbf{(pp)}}\\
\midrule
Qwen3.5-4B~\citep{qwen35blog} & 19.84 & 76.82 & 49.04
& \PASpatialCI{$-27.79$}{-32.16,-23.43}\\
\addlinespace[2pt]
Qwen3.5-9B~\citep{qwen35blog} & 19.84 & 76.82 & 51.78
& \PASpatialCI{$-25.04$}{-29.12,-20.81}\\
\addlinespace[2pt]
Qwen3.5-27B~\citep{qwen35blog} & 19.84 & 76.82 & \textbf{86.35}
& \PASpatialCI{\textbf{+9.53}}{6.10,13.06}\\
\addlinespace[2pt]
InternVL3.5-8B~\citep{wang2025internvl3} & 19.84 & 76.82 & 30.30
& \PASpatialCI{$-46.52$}{-50.58,-42.20}\\
\bottomrule
\end{tabularx}
\par\smallskip
\begin{minipage}{\linewidth}
\footnotesize\raggedright
Bold highlights the model exceeding the tested rule on matched support.
Values and differences are rounded independently. These subset scores
are not overall $GAcc$ on $\mathcal G$.
\end{minipage}
\end{table*}

Table~\ref{tab:spatial_same_support} gives the paired intervals for
the four model-minus-nearest contrasts in Section~\ref{sec:validity}.
On $\mathcal H_{\mathrm{hard}}$, the nearest-region rule scores zero
by construction. Table~\ref{tab:hard_subset_diagnostics}(a) therefore
reports each model's margin over the random-candidate expectation
computed on the same hard probes.

\begin{table*}[t]
\centering
\caption{Diagnostics on the nearest-region failure subset.
All shares and accuracies are percentages, differences are percentage
points. Brackets give 95\% source-parent-cluster bootstrap intervals.}
\label{tab:hard_subset_diagnostics}
\small
\setlength{\tabcolsep}{6pt}
\renewcommand{\arraystretch}{1.15}
\def\PAHardCI#1#2{\makecell[c]{#1\\[-1pt]{\scriptsize$[#2]$}}}
\begin{tabularx}{\linewidth}{@{}>{\raggedright\arraybackslash}p{0.20\linewidth} *{3}{>{\centering\arraybackslash}X}@{}}
\toprule
\multicolumn{4}{@{}l@{}}{\textbf{(a) Grounding on $\mathcal H_{\mathrm{hard}}$: 271 parents/394 probes}}\\
\midrule
\rowcolor{black!5}
\textbf{Model}
& \makecell[c]{\textbf{Random}\\$GAcc\,\uparrow$}
& \makecell[c]{\textbf{Model}\\$GAcc\,\uparrow$}
& \makecell[c]{\textbf{Model--Random}\\\textbf{(pp)}}\\
\midrule
Qwen3.5-4B~\citep{qwen35blog} & 18.76 & 38.03
& \PAHardCI{+19.27}{12.38,25.89}\\
\addlinespace[2pt]
Qwen3.5-9B~\citep{qwen35blog} & 18.76 & 35.21
& \PAHardCI{+16.46}{9.86,23.09}\\
\addlinespace[2pt]
Qwen3.5-27B~\citep{qwen35blog} & 18.76 & \textbf{73.22}
& \PAHardCI{\textbf{+54.47}}{48.05,60.45}\\
\addlinespace[2pt]
InternVL3.5-8B~\citep{wang2025internvl3} & 18.76 & 17.80
& \PAHardCI{$-0.96$}{-6.28,4.56}\\
\bottomrule
\end{tabularx}
\par\medskip
\begin{tabularx}{\linewidth}{@{}>{\raggedright\arraybackslash}p{0.20\linewidth} *{2}{>{\centering\arraybackslash}X}@{}}
\toprule
\multicolumn{3}{@{}l@{}}{\textbf{(b) Reading--grounding on $\mathcal L\cap\mathcal H_{\mathrm{hard}}$: 69 parents/74 probes}}\\
\midrule
\rowcolor{black!5}
\textbf{Model} & $C^{+}G^{-}$ share
& $\mathrm{GErr}\mid C\,\downarrow$\\
\midrule
Qwen3.5-4B~\citep{qwen35blog}
& \PAHardCI{27.54}{17.42,37.88}
& \PAHardCI{47.50}{33.82,62.00}\\
\addlinespace[2pt]
Qwen3.5-9B~\citep{qwen35blog}
& \PAHardCI{44.20}{32.14,56.72}
& \PAHardCI{62.24}{49.44,75.27}\\
\addlinespace[2pt]
Qwen3.5-27B~\citep{qwen35blog}
& \PAHardCI{29.71}{20.14,39.84}
& \PAHardCI{34.75}{23.77,46.55}\\
\addlinespace[2pt]
InternVL3.5-8B~\citep{wang2025internvl3}
& \PAHardCI{2.17}{0.00,5.88}
& \makecell[c]{100.00\\[-1pt]{\scriptsize---}}\\
\bottomrule
\end{tabularx}
\par\smallskip
\begin{minipage}{\linewidth}
\footnotesize\raggedright
Nearest-region $GAcc$ is zero on $\mathcal H_{\mathrm{hard}}$ by definition.
Panel (b) uses only its joint-reading intersection, not all 394 probes.
The dash denotes an unreported interval because the conditional statistic
is undefined in some resamples, it does not denote zero uncertainty.
Panel (a) bold highlights the strongest margin over random selection.
Conditional rates are diagnostic, not standalone cross-model rankings.
\end{minipage}
\end{table*}

Reading--grounding disagreement persists where the rule fails.
Both the $C^{+}G^{-}$ share and $\mathrm{GErr}\mid C$ in
Table~\ref{tab:hard_subset_diagnostics}(b) use the same
joint support $\mathcal I_m$ and weights:
\begin{equation}
\begin{aligned}
q_{10,m}^{\mathrm{hard}}
&=\operatorname{Agg}_{\mathcal I_m}\!\left(C_m(1-G_m)\right),\\
\mathrm{GErr}_{m\mid C}^{\mathrm{hard}}
&=\frac{q_{10,m}^{\mathrm{hard}}}
{\operatorname{Agg}_{\mathcal I_m}(C_m)}.
\end{aligned}
\label{eq:hard_reading_grounding}
\end{equation}
For Qwen3.5-27B~\citep{qwen35blog}, these quantities are 29.71\% [20.14, 39.84] and
34.75\% [23.77, 46.55], respectively. The first is a weighted event
share over the intersection, whereas the second conditions on correct
reading. InternVL3.5-8B's~\citep{wang2025internvl3} small event share does not imply reliable
grounding: its conditional point estimate is 100\%, and the reading
mass vanishes in some resamples. The analysis establishes that the
observed recognition--grounding discrepancy persists on this rule's
failure subset, without claiming that all geometric shortcuts have
been excluded.



\paragraph{No-Image and Candidate-Order Controls.}

The no-image control sends no image attachment while retaining the
original problem text, evidence coordinates, and candidate geometry.
It tests whether grounding performance depends on image pixels given
the remaining non-pixel cues. The candidate-order control permutes only
the candidate array, candidate aliases remain bound to the same
physical regions. This intervention tests whether a fixed presentation
order is sufficient to explain aggregate performance.

\definecolor{PADTwoBestBg}{HTML}{E8F3EC}

\definecolor{PADTwoSecondBg}{HTML}{FFF4D6}

\definecolor{PADTwoShiftBg}{HTML}{FBE9E8}


\providecommand{\PADTwoBest}[1]{#1}
\providecommand{\PADTwoSecond}[1]{#1}
\providecommand{\PADTwoShift}[1]{#1}

\renewcommand{\PADTwoBest}[1]{%
    \begingroup
    \setlength{\fboxsep}{0.8pt}%
    \colorbox{PADTwoBestBg}{%
        \textbf{#1}%
    }%
    \endgroup
}

\renewcommand{\PADTwoSecond}[1]{%
    \begingroup
    \setlength{\fboxsep}{0.8pt}%
    \colorbox{PADTwoSecondBg}{%
        #1%
    }%
    \endgroup
}

\renewcommand{\PADTwoShift}[1]{%
    \begingroup
    \setlength{\fboxsep}{0.8pt}%
    \colorbox{PADTwoShiftBg}{%
        #1%
    }%
    \endgroup
}

\begin{table*}[t]
\centering
\caption{Grounding under input controls (\%), changes are in percentage points relative to the reference condition.}
\label{tab:input_controls}
\begingroup
\small
\setlength{\tabcolsep}{3.2pt}
\renewcommand{\arraystretch}{1.12}
\begin{tabular*}{\linewidth}{@{\extracolsep{\fill}}lrrrrr@{}}
\toprule
\rowcolor{PAHeader}
\textbf{Model} & \textbf{Reference} & \textbf{No image} & $\Delta G$ (pp) & \textbf{Reordered} & $\Delta G$ (pp) \\
\midrule
Qwen3.5-4B~\citep{qwen35blog} & 40.21 & 38.78 & -1.43 & 41.93 & +1.72 \\
Qwen3.5-9B~\citep{qwen35blog} & 40.52 & 43.53 & +3.01 & 43.02 & +2.50 \\
Qwen3.5-27B~\citep{qwen35blog} & 84.11 & 63.72 & -20.39 & 84.48 & +0.37 \\
InternVL3.5-8B~\citep{qwen35blog} & 23.52 & 49.87 & +26.35 & 7.24 & -16.28 \\
GPT-6 Astra~\citep{openai2026gpt6astra} & 81.94 & 75.57 & -6.37 & 84.27 & +2.33 \\
Gemini-3.8-flash~\citep{googledeepmind2026gemini38flash} & \PABest{90.71} & \PABest{76.90} & -13.81 & \PABest{90.82} & +0.11 \\
Gemma4-26B-A4B~\citep{team2026gemma} & 69.27 & 59.83 & -9.44 & 74.59 & +5.32 \\
GLM46V-flash~\citep{vteam2025glm45vglm41vthinkingversatilemultimodal} & 54.05 & 47.47 & -6.58 & 52.49 & -1.56 \\
Kimi-VL-A3B~\citep{team2025kimi} & 34.14 & 35.45 & +1.31 & 32.79 & -1.35 \\
Molmo2-8B~\citep{team2025kimi} & 33.79 & 30.43 & -3.36 & 27.69 & -6.10 \\
\bottomrule
\end{tabular*}
\par\smallskip
\begin{minipage}{\linewidth}
\footnotesize\raggedright
Blue bold marks the maximum available accuracy in each condition.
Signed changes are descriptive and do not indicate significance.
Reference, no-image, and reordered scores are the same as in Table~\ref{tab:validity_controls}.
\end{minipage}
\endgroup
\end{table*}

Removing the image lowers grounding accuracy for seven of the ten
models. Qwen3.5-27B~\citep{qwen35blog}, GPT-6 Astra~\citep{openai2026gpt6astra}, and Gemini-3.8-flash~\citep{googledeepmind2026gemini38flash} lose 20.39,
6.37, and 13.81 percentage points, respectively. Gemma4-26B-A4B~\citep{team2026gemma},
GLM46V-flash~\citep{vteam2025glm45vglm41vthinkingversatilemultimodal}, and Molmo2-8B~\citep{clark2026molmo2} lose 9.44, 6.58, and 3.36 points,
while Qwen3.5-4B~\citep{qwen35blog} changes by $-1.43$ points.
Qwen3.5-9B~\citep{qwen35blog}, InternVL3.5-8B~\citep{wang2025internvl3}, and Kimi-VL-A3B~\citep{team2025kimi} instead gain 3.01,
26.35, and 1.31 points. These divergent responses show that access
to an image does not uniformly improve grounding.
Because removing the image can affect evidence anchors and target
identifiability, the control measures sensitivity to image input,
not the isolated causal contribution of visual reasoning.

Candidate reordering has model-dependent effects.
Seven models change by at most 2.50 percentage points in absolute
$GAcc$. InternVL3.5-8B~\citep{wang2025internvl3} falls from 23.52\% to 7.24\% ($-16.28$ points),
Gemma4-26B-A4B~\citep{team2026gemma} rises from 69.27\% to 74.59\% ($+5.32$), and
Molmo2-8B~\citep{clark2026molmo2} falls from 33.79\% to 27.69\% ($-6.10$).
Small aggregate changes for the other seven models
do not establish permutation invariance or rule out candidate-order
effects: gains and losses on individual probes can offset one another.

The controls support distinct, limited conclusions. Several models
lose accuracy without images, the tested proximity rule does not
fully explain Qwen3.5-27B's~\citep{qwen35blog} grounding performance, and candidate
reordering affects models differently. These results constrain
explanations based on image input, geometry, and presentation without
identifying internal strategies or excluding untested shortcuts.

\subsection{Grounding and Original-Problem Solving}
\label{app:problem_solving}

This section documents the grouping behind
Section~\ref{sec:association} and Figure~\ref{fig:association}.
Matching links each model's original-problem outcomes to its local-probe
records from the full release. For the three Qwen3.5 models,
InternVL3.5-8B~\citep{wang2025internvl3}, GPT-6 Astra~\citep{openai2026gpt6astra}, and Gemini-3.8-flash~\citep{googledeepmind2026gemini38flash}, the recorded matched
analysis contains 311 parents and 412 probes. These counts describe
that six-model diagnostic, not the denominator of $SolveAcc$, which
uses the 986-problem release. Matched coverage is distinct from
release eligibility and is not inferred for other models from equal
aggregate scores.
Solving and probing use independent contexts, without shared
history, predictions, or scoring feedback.

The solve-correct and solve-wrong groups refer to the original
binary final-answer correctness outcomes. These labels are distinct
from the normalized partial-credit scores $s_i$ in
Appendix~\ref{app:solve_scoring}, $SolveAcc$ is not the proportion
of parents in the solve-correct group.
For each model, local grounding is averaged within each parent
and then across parents in each binary-outcome group, yielding
$\bar G_{\mathrm{wrong}}$ and $\bar G_{\mathrm{correct}}$.
We report $\Delta_{\mathrm{solve}}=
\bar G_{\mathrm{correct}}-\bar G_{\mathrm{wrong}}$.
Table~\ref{tab:app-solving-association} retains these
outcome-conditioned estimates.

\begin{table*}[t]
\centering
\caption{Parent-level grounding by binary final-answer correctness (\%), distinct from partial-credit $SolveAcc$.}
\label{tab:app-solving-association}
\begingroup
\small
\setlength{\tabcolsep}{3.2pt}
\renewcommand{\arraystretch}{1.12}
\begin{tabular*}{\linewidth}{@{\extracolsep{\fill}}lrrr@{}}
\toprule
\rowcolor{PAHeader}
\textbf{Model} & $\bar G_{\mathrm{wrong}}$ & $\bar G_{\mathrm{correct}}$ & $\Delta_{\mathrm{solve}}$ (pp) \\
\midrule
Qwen3.5-4B~\citep{qwen35blog} & 39.19 & 41.31 & +2.12 \\
Qwen3.5-9B~\citep{qwen35blog} & 40.67 & 44.61 & +3.94 \\
Qwen3.5-27B~\citep{qwen35blog} & 85.51 & 87.89 & +2.39 \\
InternVL3.5-8B~\citep{wang2025internvl3} & 23.37 & 25.88 & +2.51 \\
GPT-6 Astra~\citep{openai2026gpt6astra} & 83.05 & 85.53 & +2.48 \\
Gemini-3.8-flash~\citep{googledeepmind2026gemini38flash} & 92.08 & 92.36 & +0.28 \\
Gemma4-26B-A4B~\citep{team2026gemma} & 50.00 & 70.95 & +20.95 \\
GLM46V-flash~\citep{thrush2022winoground} & 58.33 & 63.98 & +5.65 \\
Kimi-VL-A3B~\citep{team2025kimi} & 33.89 & 36.63 & +2.74 \\
Molmo2-8B~\citep{clark2026molmo2} & 32.67 & 36.22 & +3.55 \\
\bottomrule
\end{tabular*}
\par\smallskip
\begin{minipage}{\linewidth}
\footnotesize\raggedright
Differences retain the supplied estimates, rounded independently of the displayed group means.
Each model uses its own matched records and binary final-answer groups, group membership is not shared across models.
These are descriptive associations, not causal effects.
\end{minipage}
\endgroup
\end{table*}

All ten correct-minus-wrong differences are positive point estimates.
Gemma4-26B-A4B~\citep{team2026gemma} has the largest gap (20.95 points), the other nine
range from 0.28 to 5.65 points. The association therefore has a
consistent direction but a model-dependent magnitude.
Correct answers still coexist with local grounding errors: the
solve-correct grounding accuracies of Qwen3.5-27B~\citep{qwen35blog} and GPT-6 Astra~\citep{openai2026gpt6astra} are
87.89\% and 85.53\%, respectively. Conversely, Gemini-3.8-flash~\citep{googledeepmind2026gemini38flash}
retains 92.08\% grounding accuracy in its solve-wrong group.

These outcome-conditioned means use parent-level averaging and need not equal
the type-balanced $GAcc$ in Table~\ref{tab:main_results}. Group membership is
determined separately for each model and may reflect problem difficulty and
other unobserved factors, the solve-correct and solve-wrong subsets are not
fixed across models. The reported differences are descriptive point estimates,
which alone establish neither statistical significance nor causality.
Independent probing also cannot establish which local evidence was used
during solving or whether correcting a mismatch would alter the final answer.
The cases in Appendix~\ref{app:casestudy} illustrate these observable
distinctions without estimating their prevalence.


\subsection{Representative Case Studies}
\label{app:casestudy}

Each case presents the source evidence, candidate entities, and Base
and +GT probe outputs. Cases~1--3 also report binary final-answer
outcomes from independent solving runs. Case~4 provides the recorded
grounding response and the controlled substitution discussed in
Section~\ref{sec:association}. The four cases illustrate individual
behaviors. Their selection does not estimate error frequencies.

\begin{figure*}[t]
    \centering
    \includegraphics[width=\textwidth]{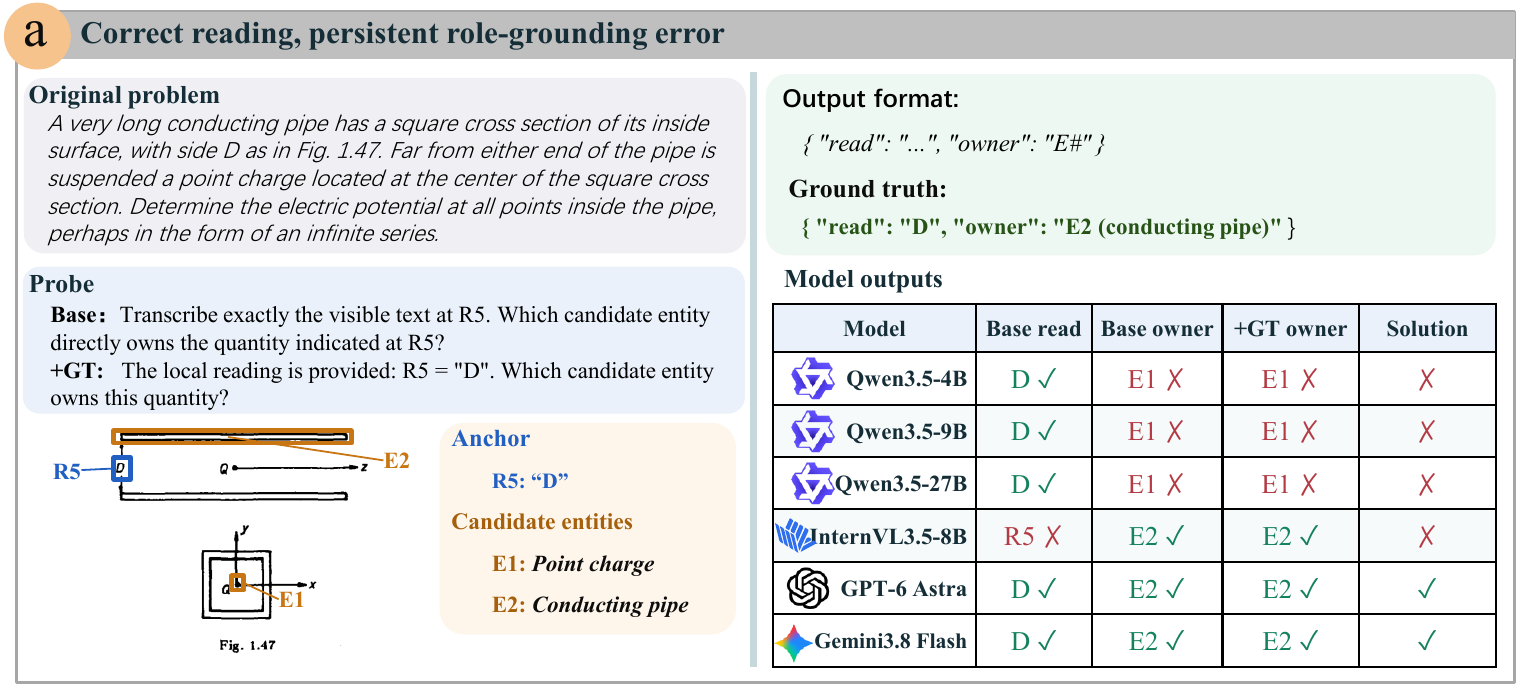}
    \caption{Case 1: A role mismatch persists despite correct reading. All
    three Qwen3.5 models correctly read $D$ at anchor R5, yet assign it to the
    point charge E1 rather than the conducting pipe E2 under both Base and
    +GT.}
    \label{fig:case-persistent-mismatch}
\end{figure*}

\begin{figure*}[t]
    \centering
    \includegraphics[width=\textwidth]{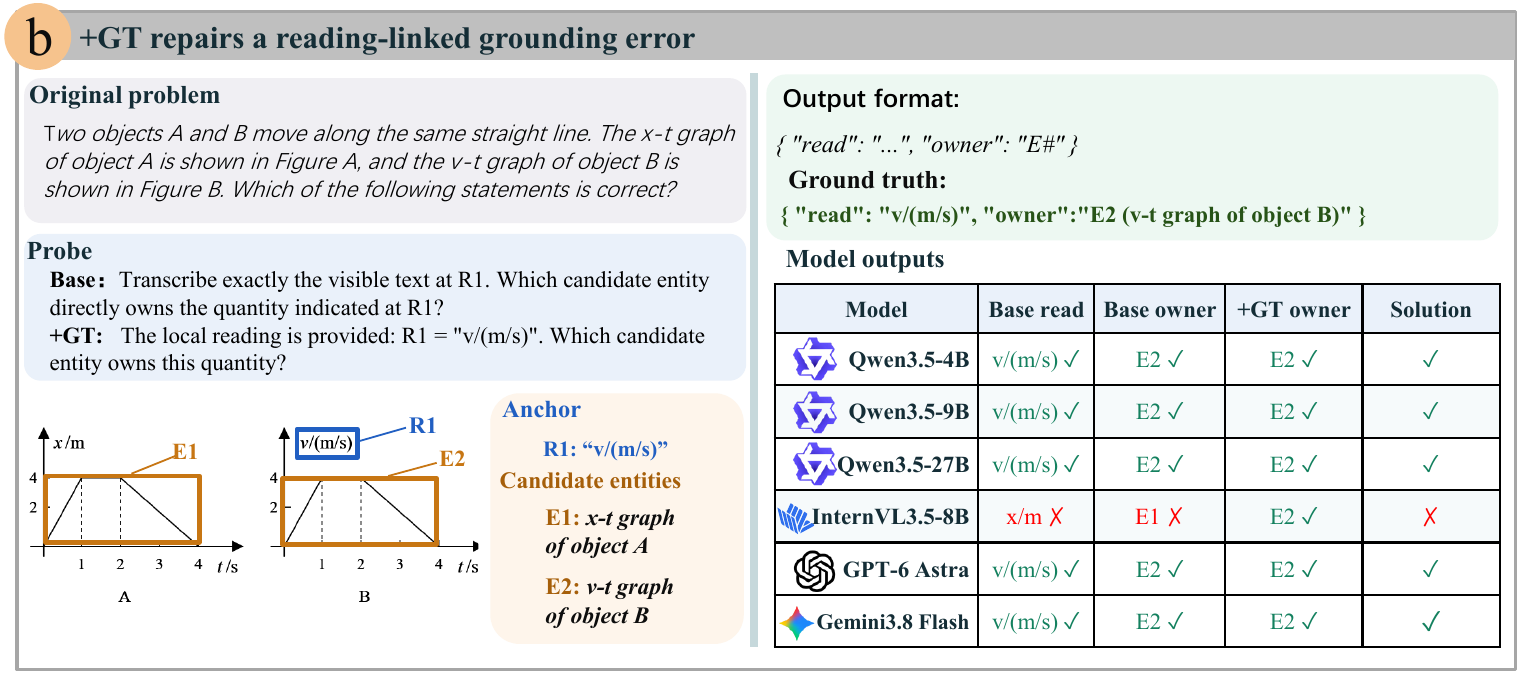}
    \caption{Case 2: A grounding error is repaired under +GT.
    Under Base, InternVL3.5-8B~\citep{wang2025internvl3} misreads the vertical-axis label and selects E1,
    after receiving the correct reading, it changes its owner prediction to
    E2.}
    \label{fig:case-reading-mediated-repair}
\end{figure*}

\begin{figure*}[t]
    \centering
    \includegraphics[width=\textwidth]{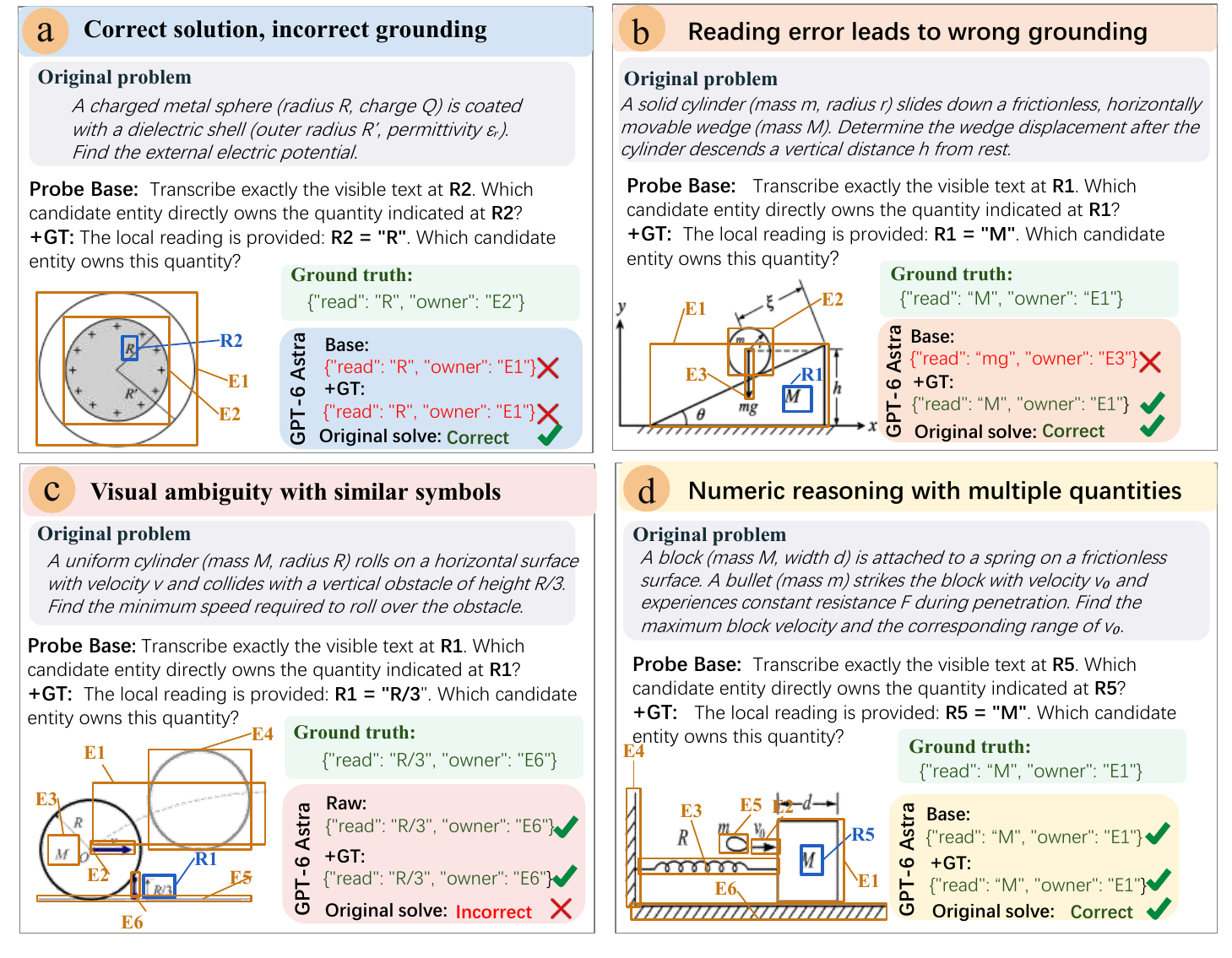}
    \caption{Case 3: Local grounding and original-problem solving are not
    equivalent measures. Panels (a) and (c) show a correct solution with
    incorrect grounding and correct grounding with an incorrect solution,
    respectively, panel (d) shows a consistent success.}
    \label{fig:case-grounding-solving-combinations}
\end{figure*}

\begin{figure*}[t]
    \centering
    \includegraphics[width=\textwidth]{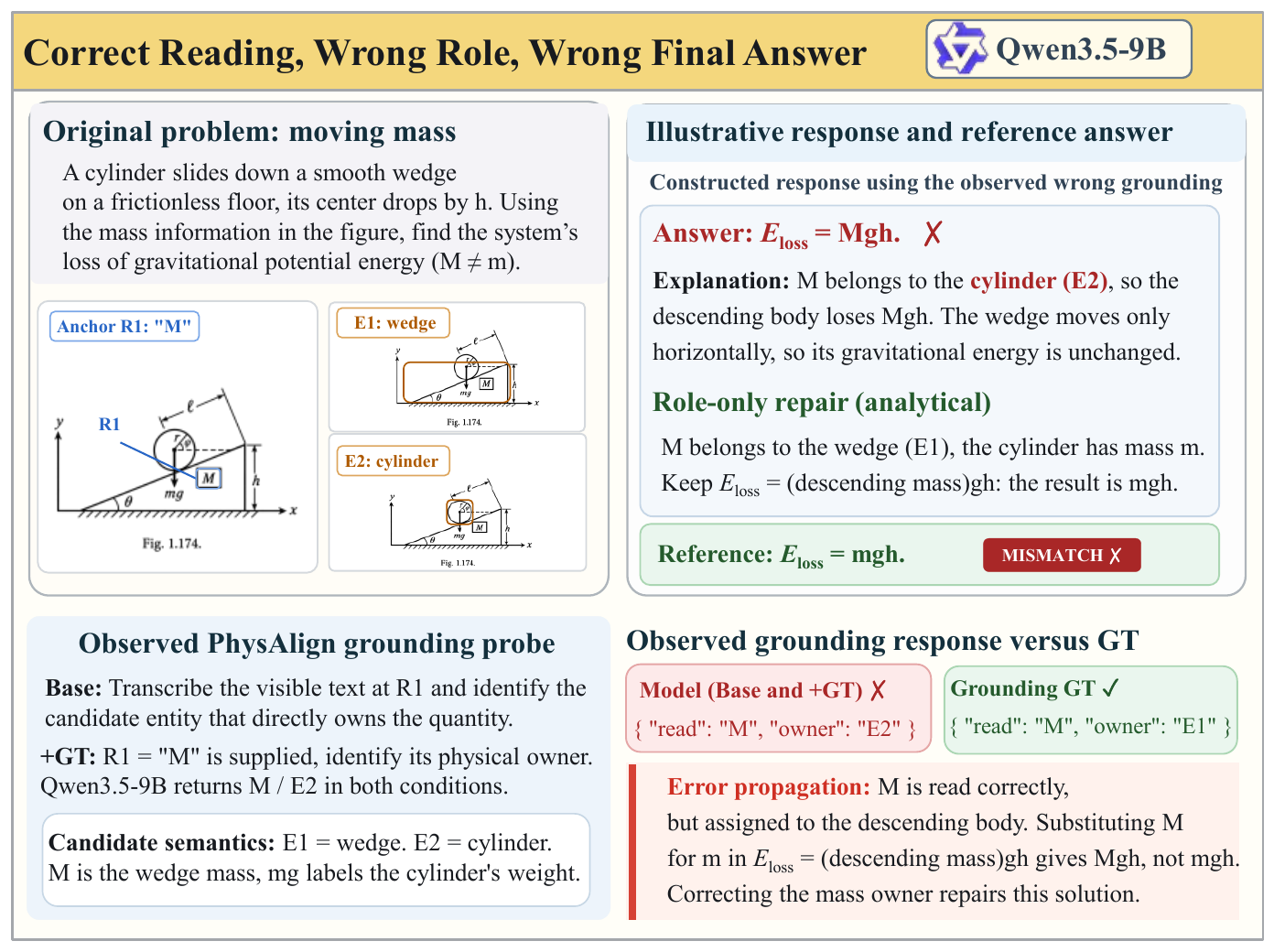}
    \caption{Case 4. The lower panels show the query, the recorded
    Qwen3.5-9B~\citep{qwen35blog} response under Base and +GT, and the
    reference assignment. The upper-right panel shows the authors'
    analytical substitution for each assignment.}
    \label{fig:case-controlled-role-substitution}
\end{figure*}

Figure~\ref{fig:case-persistent-mismatch} directly instantiates the
$C^{+}G^{-}$ pattern discussed in the main text. Qwen3.5-4B~\citep{qwen35blog}, Qwen3.5-9B~\citep{qwen35blog}, and
Qwen3.5-27B~\citep{qwen35blog} all transcribe the side-length symbol $D$ correctly but select the
wrong physical owner, E1. Their predictions remain unchanged when +GT
explicitly supplies the correct reading. The failure therefore lies beyond
local transcription, in the association between the recognized content and
its physical entity. GPT-6 Astra~\citep{openai2026gpt6astra} and Gemini-3.8-flash~\citep{googledeepmind2026gemini38flash} recover both the reading and the owner from
the same evidence, providing evidence that the intended relation is
recoverable from the input. Correct recognition alone does not guarantee
correct physical-role grounding.

Figure~\ref{fig:case-reading-mediated-repair} shows a complementary
paired outcome. Anchor R1 marks the vertical-axis label
$v/(\mathrm{m}/\mathrm{s})$ of object B's $v$--$t$ graph, with owner E2.
Under Base, InternVL3.5-8B~\citep{wang2025internvl3} reports
$x/\mathrm{m}$ and selects E1, under +GT, its owner prediction changes
to E2. This transition is consistent with a reading-linked error,
but does not establish causal mediation by recognition.
Together with Figure~\ref{fig:case-persistent-mismatch}, it illustrates
that supplied readings can accompany either repaired or persistent
role errors. The independent solving response is incorrect, because
that run did not receive the +GT input, this case does not test
whether the added reading would improve problem solving.

Figure~\ref{fig:case-grounding-solving-combinations} establishes the
bidirectional non-equivalence between the local diagnostic and
original-problem solving. In panel~(a), GPT-6 Astra~\citep{openai2026gpt6astra} solves the original problem
correctly but consistently assigns $R'$ to the wrong entity E1 under both Base
and +GT. A correct final answer thus does not imply consistent recovery of
local physical roles. Panel~(c) exhibits the converse: the model correctly
reads and grounds $R/3$, yet answers the original problem incorrectly, showing
that one correct grounding decision cannot rule out subsequent modeling or
derivation errors. Panel~(b) provides another instance in which +GT repairs a
reading-mediated error, whereas panel~(d) shows correct recognition, grounding,
and solving in a scene containing several annotated quantities. The probe
therefore distinguishes consistent success from reading errors, role
mismatches, and downstream reasoning failures rather than merely introducing
additional opportunities for failure.

In Figure~\ref{fig:case-controlled-role-substitution}, the
\texttt{read} and \texttt{owner} fields at anchor R1 are $(M,\mathrm{E2})$
under both inputs. The reference pair is $(M,\mathrm{E1})$. E1 denotes
the wedge and E2 the cylinder. These are the two assignments used
for the energy comparison in Section~\ref{sec:association}.

\FloatBarrier
\section{Comparison with Related Benchmarks}
\label{app:benchmark_comparison}

We extend Section~\ref{sec:related_work} by distinguishing problem-level
reasoning, structured visual understanding, and controlled error diagnosis.

\paragraph{Problem- and step-level reasoning.}
MathVista~\citep{lu2024mathvista} evaluates mathematical reasoning in visual
contexts, and PhysUniBench~\citep{wang2025physunibench} focuses on
undergraduate physics. PhysReason~\citep{zhang2025physreason} evaluates both
final answers and intermediate solution steps. PhysAlign instead tests local
content--role correspondences, without requiring complete solutions or
treating solution-derived facts as probe targets.

\paragraph{Diagram structure, relations, and reading.}
AI2D~\citep{hiippala2021ai2d} represents diagram constituents and relations
as diagram parse graphs. GQA~\citep{hudson2019gqa} constructs compositional
questions from scene graphs and measures consistency and grounding.
TextVQA~\citep{singh2019textvqa} requires reading and reasoning about image
text. PhysAlign likewise emphasizes structured visual evidence, but provides
fixed anchors and candidates and scores literal recognition separately from
physical-role correspondence. It neither requires complete diagram parsing
nor equates role grounding with visual localization.

\paragraph{Controlled error diagnosis.}
HallusionBench~\citep{guan2024hallusionbench} uses controlled image--question
groups to diagnose language hallucination and visual illusion. PhysAlign
tests whether supplying a verified local reading changes the reported role,
keeping the original problem, anchor, candidates, and scoring fixed. This
control neither removes all perceptual uncertainty nor reveals internal
reasoning.

Table~\ref{tab:related_benchmarks} compares the protocols of selected closely
related benchmarks. Its criteria concern diagnostic interfaces, not overall
benchmark quality. Associations between PhysAlign scores and independently
evaluated problem solving are descriptive, not causal.

\section{Scope, Limitations, and Release Documentation}
\label{app:appendix_limitations}

\paragraph{Measurement Scope.}
PhysAlign measures whether a model can map locally visible content to its
correct physical role when given evidence anchors, candidate entities, and a
constrained output interface. The protocol scores content recognition and
role grounding separately, but it does not assess unconstrained, holistic
diagram understanding. The +GT condition supplies only a verified local
transcription: it reveals neither the correct candidate nor the role label,
and it leaves uncertainty in entity boundaries, spatial relations, and
physical semantics unresolved. The Base/+GT comparison therefore measures
the change in grounding after a correct transcription is supplied, it is not
an upper bound under perfect perception.

\paragraph{Interpretive Boundaries.}
Our measurements concern explicit local correspondences reported under fixed
inputs. They do not recover a model's latent representation, complete
physical world model, or internal reasoning process. Local probes and
original-problem solving are evaluated in separate contexts. Agreement or
disagreement between them supports a distinction at the level of observable
behavior, but cannot determine which local premise the model used, ignored,
or propagated incorrectly in a particular solution. Likewise, a controlled
substitution based on an incorrect role characterizes a possible consequence
within a fixed derivation, it does not causally identify the model's actual
reasoning trace.

\paragraph{Coverage and Release Boundaries.}
The current evaluation focuses on static educational physics diagrams and is
skewed toward mechanics and electromagnetism. Retained probes cover reviewed
local facts rather than every expressible relation in each problem, and
paired +GT inputs are available only for a subset. Because the parent
problems come from public datasets, we cannot exclude prior exposure to the
original problems or semantically similar variants during pretraining.
Parent-level split propagation keeps probes from one parent together,
and exact-input deduplication removes identical public inputs. Neither
certifies a contamination-free evaluation. The reported experiments
combine development and test within the retained release and should
therefore be interpreted as release-derived diagnostics, not held-out
test estimates. Appendix~\ref{app:dataset} documents
provenance, review, retention, and coverage.
These documentation priorities are consistent with
Datasheets for Datasets~\citep{gebru2021datasheets},
which advocates explicit reporting of dataset composition,
collection, and recommended uses. We plan to release the probe
inputs and metadata, prompt templates, evaluation code, and recorded run
configurations for which redistribution is permitted, while keeping public
inputs separate from private reference targets and scoring mappings. Source
problems and images will be redistributed only where upstream licenses
permit, otherwise, the release will provide provenance identifiers and
reconstruction instructions without asserting additional rights over
third-party content.

\par\medskip
\begingroup
\ifdefined\nolinenumbers\nolinenumbers\fi
\noindent\begin{minipage}{\textwidth}
    \captionsetup{type=table}
    \centering

    \renewcommand{\PAbench}[2]{\mbox{#1~\citep{#2}}}

    \small
    \setlength{\tabcolsep}{4pt}
    \renewcommand{\arraystretch}{1.16}

    \renewcommand{\tabularxcolumn}[1]{m{#1}}

    \begin{threeparttable}

        \caption{Protocol-level comparison with closely related benchmarks.}
        \label{tab:related_benchmarks}

        \begin{tabularx}{\linewidth}{
            @{}
            >{\raggedright\arraybackslash}m{0.31\linewidth}
            >{\hsize=0.99\hsize\linewidth=\hsize\centering\arraybackslash}X
            >{\hsize=0.99\hsize\linewidth=\hsize\centering\arraybackslash}X
            >{\hsize=0.86\hsize\linewidth=\hsize\centering\arraybackslash}X
            >{\hsize=1.09\hsize\linewidth=\hsize\centering\arraybackslash}X
            >{\hsize=1.10\hsize\linewidth=\hsize\centering\arraybackslash}X
            >{\hsize=1.17\hsize\linewidth=\hsize\centering\arraybackslash}X
            >{\hsize=0.80\hsize\linewidth=\hsize\centering\arraybackslash}X
            @{}
        }
            \toprule

            \multirow[c]{2}{*}{\textbf{Benchmark}}
            &
            \multicolumn{2}{c}{\textbf{General evaluation}}
            &
            \multicolumn{4}{c}{\textbf{Localized diagnosis}}
            &
            \textbf{Utility}
            \\

            \cmidrule(lr){2-3}
            \cmidrule(lr){4-7}
            \cmidrule(l){8-8}

            &
            \shortstack[c]{\textbf{Input}\\\textbf{Ctrl.}}
            &
            \shortstack[c]{\textbf{Interm.}\\\textbf{Eval.}}
            &
            \shortstack[c]{\textbf{Fixed}\\\textbf{Anchor}}
            &
            \shortstack[c]{\textbf{Rec./Grd.}\\\textbf{Split}}
            &
            \shortstack[c]{\textbf{Paired}\\\textbf{Rec. Ctrl.}}
            &
            \shortstack[c]{\textbf{Det. Local}\\\textbf{Score}}
            &
            \shortstack[c]{\textbf{Solve}\\\textbf{Link}}
            \\[-0.5pt]

            \midrule

            \PAbench{SeePhys}{xiang2026seephys}
            & \PAyes & \PAno & \PAno & \PAno & \PAno & \PAno & \PAno \\
            \addlinespace[2pt]

            \PAbench{SeePhys Pro}{xiang2026seephyspro}
            & \PAyes & \PApart & \PAno & \PAno & \PAno & \PAno & \PAno \\
            \addlinespace[2pt]

            \PAbench{PhysicsArena}{dai2025physicsarena}
            & \PAno & \PAyes & \PAno & \PAno & \PAno & \PAno & \PAyes \\
            \addlinespace[2pt]

            \PAbench{MathVerse}{zhang2024mathverse}
            & \PAyes & \PAyes & \PAno & \PAno & \PAno & \PAno & \PAno \\

            \midrule

            \rowcolor{black!8}
            \textbf{PhysAlign (Ours)}
            & \PAyes & \PAyes & \PAyes & \PAyes & \PAyes & \PAyes & \PAyes \\

            \bottomrule
        \end{tabularx}

        \begin{tablenotes}[flushleft]
            \footnotesize
            \item
            \textbf{Legend:}
            \PAyes{} explicit,
            \PApart{} indirect or related,
            \PAno{} no corresponding protocol under the definitions below.
            Symbols indicate protocol features, not model performance or
            overall benchmark quality.
            \textbf{Input Ctrl.}: controlled changes to evidence
            availability or modality.
            \textbf{Interm. Eval.}: evaluation of intermediate reasoning
            stages or outputs.
            \textbf{Fixed Anchor}: evaluation at a predefined local image
            region or text span in an unchanged source problem.
            \textbf{Rec./Grd. Split}: separate recognition and
            role-grounding scores on the same probe.
            \textbf{Paired Rec. Ctrl.}: matched probes with and without
            correct local readings, with the grounding target withheld.
            \textbf{Det. Local Score}: rule-based local scoring without
            an LLM judge.
            \textbf{Solve Link}: association between separately evaluated
            diagnostic scores and original-problem correctness.
            For SeePhys Pro, \PApart{} denotes modality-transfer-based
            diagnosis rather than direct intermediate-output scoring.
        \end{tablenotes}

    \end{threeparttable}
\end{minipage}\par
\endgroup

\FloatBarrier
{
  \small

}

\end{document}